\documentclass[journal,comsoc]{IEEEtran}

\usepackage{booktabs}

\usepackage{hyperref}

\usepackage[utf8]{inputenc}
\usepackage{amssymb}
\usepackage{amsmath}

\usepackage{tabularx}
\usepackage{mathtools}
\usepackage{graphicx}
\usepackage{adjustbox}
\usepackage{algorithm}
\usepackage{algorithmic}
\usepackage[FIGTOPCAP]{subfigure}
\usepackage{multirow}
\usepackage{booktabs}
\usepackage{enumitem}
\usepackage{color}

\DeclarePairedDelimiter\ceil{\lceil}{\rceil}

\graphicspath{/figs}

\begin{document}
	
	\title{Random Partitioning Forest for Point-Wise and Collective Anomaly Detection - Application to Network Intrusion Detection}
	
	
	\author{Pierre-Francois Marteau,~\IEEEmembership{Member,~IEEE,}
		\thanks{P-F. Marteau is with the IRISA UMR CNRS 6074,
			Université Bretagne Sud,
			Campus de Tohannic, 56000 Vannes, France, e-mail: (see http://https://people.irisa.fr/Pierre-Francois.Marteau/).}
		\thanks{Manuscript received April 19, 2005; revised August 26, 2015.}}
	
	\markboth{Journal of \LaTeX\ Class Files,~Vol.~14, No.~8, August~2015}%
	{Shell \MakeLowercase{\textit{et al.}}: Bare Demo of IEEEtran.cls for IEEE Communications Society Journals}
	
	
	
	
	
	\maketitle

\begin{abstract}
In this paper, we propose DiFF-RF, an ensemble approach composed of random partitioning binary trees to detect point-wise and collective (as well as contextual) anomalies. 
Thanks to a distance-based paradigm used at the leaves of the trees, this semi-supervised approach solves a drawback that has been identified in the isolation forest (IF) algorithm. Moreover, taking into account the frequencies of visits in the leaves of the random trees allows to significantly improve the performance of DiFF-RF when considering the presence of collective anomalies. DiFF-RF is fairly easy to train, and good performance can be obtained by using a simple semi-supervised procedure to setup the extra hyper-parameter that is introduced. We first evaluate DiFF-RF on a synthetic data set to i) verify that the limitation of the IF algorithm is overcome, ii) demonstrate how collective anomalies are actually detected and iii) to analyze the effect of the meta-parameters it involves. We assess the DiFF-RF algorithm on a large set of datasets from the UCI repository, as well as four benchmarks related to network intrusion detection applications. Our experiments show that DiFF-RF almost systematically outperforms the IF algorithm and one of its extended variant, but also challenges the one-class SVM baseline, deep learning variational auto-encoder and ensemble of auto-encoder architectures. Finally, DiFF-RF is computationally efficient and can be easily parallelized on multi-core architectures. 
\end{abstract}


\section{Introduction}

Recent reviews on IDS are proposing a stable typology of the state of the art approaches \cite{Agrawal2015, Fitriani2016, Khraisat2019}. 
According to these studies, apart from state protocol analysis (SPA) approaches that rely on equipment vendor profiles, the detection method in IDS can be divided into two main categories: i) anomaly-based or ii) signature-based. These two categories are complementary, as the latter is very effective in detecting known attacks, while anomaly-based approaches are capable of detecting "zero-day" attacks, i.e. malicious events that are neither observed nor reported yet. We focus in this paper on semi-supervised anomaly-based approaches for IDS.

Anomaly detection has been a hot topic for several decades and has led to numerous applications in a wide range of domains, such as fault tolerance in industry, crisis detection in finance and economy, health diagnosis, extreme phenomena in earth science and meteorology, atypical celestial object detection in astronomy or astrophysics, system intrusion in cyber-security, etc. 

Anomaly detection is generally defined as the problem of identifying patterns that deviate from a  'normality' behavioral model, namely a model that is fitted from known normal data only. According to this definition, anomaly detection falls into the semi-supervised learning framework, a broad machine learning area in which our work is positioned. 

In the context of intrusion detection, semi-supervised approaches to anomaly detection are of great importance due to their ability to detect zero-day attacks that are in general not detected by signature based approaches.

In the literature, most semi-supervised anomaly detection approaches can be categorized either according to the model of normality that is involved or to the way they address the abnormality characterization and its identification. 

A quite exhaustive, although a bit dated,  review in anomaly detection has been proposed in \cite{Chandola:2009}, completed by a more recent comparative study \cite{Goldstein2016}. According to these studies, the state of the art methods can be distributed into five main categories:

\begin{enumerate}
	\item \textbf{Near neighbors and clustering based methods} \cite{Mennatallah2012}: Near Neighbors methods rely on the assumption that a 'normal' instance occurs close to its near neighbors while an anomaly occurs far from its near neighbors. Similarly, cluster based methods rely generally on the assumption that a 'normal' instance occurs near its closest cluster centroid while an anomaly will occur far from its nearest cluster centroid \cite{key:articleBama,Lin201513}.  However, some cluster-based methods assume that the training data may contain (unlabeled) anomalies that form their own (small and isolated) clusters. In that context, many group anomaly detection methods have been developed, one can mentioned \cite{Xiong2011} in the deep learning framework.
	\item \textbf{Classification based method}: in this paradigm, several classes of 'normal' data are learned by a set of one against all classifiers (each classifier is associated to a class and is trained to separate it from the others classes). An instance that is not categorized as 'normal' by any of these classifiers is considered as an anomaly. A peculiar case occurs when a single class is used to model the 'normal' data.  Random Forest, including recent advances on one-class random forest \cite{Desir2013}, multi-class and one-class Support Vector Machine (SVM) \cite{Fujimaki2008}, and neural networks \cite{Lee:1998:DMA:1267549.1267555,Ghosh:1999:SUN:1251421.1251433,ryan:nips10}, are the most used classifiers for anomaly detection. 
	\item \textbf{Statistical based methods} rely on the assumption that 'normal' data are associated to high probability states of an underlying stochastic process while anomalies are associated to low probability states of this process. Popular approaches in this category are kernel based density models and  the  Gaussian Mixture Model (GMM), including recent advance in one-class GMM \cite{Kemmler2013}. 
	\item \textbf{Information theoretic based methods} use  information theoretic measures \cite{Lee2001}, such as the entropy, the Kolmogorof complexity, the minimum description length, etc,  to estimate the 'complexity' of the 'normal' dataset (or equivalently the complexity of the process behind the production of these data) \cite{GFDLS2006}. If $\mathcal{C}(D)$ estimates the complexity of dataset $D$, the minimal subset of instance $I$ that maximizes $\mathcal{C}(D) - \mathcal{C}(I)$ is considered as the anomaly subset.
	\item \textbf{Spectral based method} rely on the assumption that it is possible to embed the data into a lower dimensional subspace in which 'normal' data and anomalies are supposedly well separated \cite{Akoglu2015}. PCA and graph (of similarity) clustering are among the most popular methods in this category.\\
\end{enumerate} 

We can think of a sixth class of method covering recent advances in deep learning and self-encoding based methods. These approaches have been historically initiated by Kramer \cite{Kramer1991}   and adapted recently to a deep learning framework under the form of auto-encoder (AE) \cite{Vincent2010} and Variational Auto-Encoder (VAE) \cite{Diederik2019}. In the context of anomaly detection, reconstruction error is the criterion used to decide whether a data item is normal or deviates too much from normality. The main advantage of VAE against AE is that their latent spaces are, by design, continuous, thanks to the prediction of a mean and a variance vectors allowing to smooth locally the latent space. In \cite{Mirsky2018} the authors have proposed KitNET, an online unsupervised anomaly detector based on an ensemble of autoencoders, which are trained to reconstruct the input data, and whose performance is expected to incrementally improves overtime. One particularity of KitNET is that it estimates in an unsupervised manner the number of auto-encoders in the ensemble and the dimensions of the encoding layers.  The last layer of the KitNET architecture is also an auto-encoder that takes as inputs the Root Mean Square Errors of the auto-encoders in the ensemble and provides in output the final reconstruction vector and RMSE. KitNET is considered as the state of the art unsupervised on-line anomaly detection for intrusion detection on network systems. \\


In 2008, Isolation Forest (IF) \cite{Liu2008}, a quite conceptually different approach to the previously referenced methods has been proposed. The IF paradigm is based on the \textit{difficulty} to isolate a particular instance inside the whole set of instances when using (random) partitioning tree structures. It relies on the assumption that an anomaly is in general much easier to isolate than a 'normal' data instance. Hence, IF is an unsupervised approach that relates somehow to the information theoretic based methods since the \textit{isolation difficulty} is addressed through an algorithmic complexity scheme. IF has been successfully used in some applications, in particular in the IDS context, \cite{Liu2012}, \cite{Ding2013} and has been recently extended in \cite{Aryal2014}, \cite{Shen2016}, \cite{Liao2018}, \cite{Cheng19} or \cite{Hariri2018} to improve the selection of attributes and their split values when constructing the tree. For instance, for the Extended Isolation Forest (EIF) approach \cite{Hariri2018}, a separation hyperplane with a random orientation is selected instead of selecting randomly a single dimension as the cut feature.  The main advantage of IF based algorithms is their capability to process large amount of data in high dimension spaces with computational and spatial controlled efficiency compared to other unsupervised methods. Unfortunately, as shown in \cite{Marteau2017}, IF suffers from what we call `blind spots', namely empty portions of the space in which data instances are embedded, that are nevertheless considered as normality spots by the IF algorithm. `Blind spots' effects are detailed in part II section A of the article.

The aim of this paper is to propose, a semi-supervised ensemble approach based on random partitioning trees that we refer to as DiFF-RF, suited for IDS purposes. Although DiFF-RF is essentially based on a random forest structure, just as IF, it differs fundamentally on the way anomalies are characterized. The use of a distance based criteria allows to solve the `blind spots' mis-detection of IF. Moreover, taking into account the relative frequency of visit at the leaves of the trees provides a complementary discriminant information that improves greatly the detection when facing collective anomalies. 

We detail the DiFF-RF algorithm in the second section of this paper by first introducing the distance-based and the relative frequency of visit paradigms at leaf level. We then provide some highlights about the way the so-called `blind spot' mis-detections of the IF algorithm are effectively solved by using a synthetic dataset. On these data we carried out a hyper-parameter sensibility study to estimate satisfactory ranges for default values (number of trees and sample size) and present a simple semi-supervised procedure to setup the extra parameter (distance scaling) introduced in DiFF-RF.  The third section addresses an extensive experimentation using UCI datasets from various domains and finally highlights an application in intrusion detection by exploiting two public domain benchmarks. This experimentation assesses the benefits brought by the DiFF-RF algorithm in point-wise and collective anomaly detection. Our results show that the proposed DiFF-RF algorithm compares advantageously with the state of the art baselines in anomaly detection that we have considered, namely one-class SVM, deep variational auto-encoder (VAE), ensemble of AE (KitNET), Isolation Forrest (IF) and Extended Isolation Forrest (EIF).  The fourth section develops an argument supporting the use of DiFF-RR in the context of IDS. A general discussion and some perspectives conclude the article.

\section{The DiFF-RF algorithm}

Just as the IF algorithm, DiFF-RF is nothing but a forest of binary partitioning trees. But, contrarily to the IF that uses the expected length of the path required to locate data as the anomaly score (an 'outlier' is expected to have a shorter path than an 'inlier'), DiFF-RF uses an expected distance measure to the centroid associated to the leaves of the trees to decide wether the tested data is a point-wise anomaly or not. A relative frequency of visit principle is also implemented at leaf level leading to a scoring that is aggregated to the distance score when collective anomalies are considered.

\subsubsection{Building the DiFF-RF forest}
Let $X_n \subset \mathbb{R}^d  $ be the set of training (normal) instances. 
The DiFF-RF algorithm is an ensemble based approach that builds a forest of random binary partitioning trees. Given a sample $S$ randomly drawn from $X_n$, a DiFF tree $T(S)$ is recursively built according to the (DiFF-Tree) algorithm \ref{Algo:DiFF-Tree}.

Two meta-parameters are required to build a DiFF-RF: $\psi$, the size of the subsets $S$ that are used to build the trees, and $t$, the number of trees. Parameter $h_{max}$, the maximum height of the trees, is empirically set up to $\ceil{log_2 \psi}$. 

Finally, the DiFF forest $F=\{T(S_1), T(S_2), \cdots, T(S_t)\}$ is obtained by randomly selecting $\{S_1, S_2, \cdots, S_t\}$, $t$ samples in $X_n$ with $|S_i|=\psi$ for all $i$, and constructing a DiFF-Tree on each of these samples, as depicted in Algorithm \ref{Algo:DiFF-Tree}.\\

\begin{algorithm}[]
	\begin{algorithmic}[1]
		\REQUIRE{$S \subset X_n$, $h$ the current depth level, $h_{max}$ the maximal depth limit}\\
		\ENSURE{DiFF-Tree}
		\IF{$h \ge h_{max}$ or $|S| \le 1$}
		\STATE $f_n = |S|/\psi$
		\IF{$|S| \ge 0$}
		\STATE $M_S$ $\leftarrow$ Mean($S$);
		\STATE $\sigma_S$ $\leftarrow$ StandardDeviation($S$);
		\ELSE
		\STATE $M_S$ $\leftarrow$ None
		\STATE $\sigma_S$ $\leftarrow$ None
		\ENDIF		
		\RETURN leafNode($S$, $M_S$, $\sigma_S$, $f_r$)\;
		\ELSE
		\STATE $D \leftarrow $ get\_qDistribution($S$)\;
		\STATE \textbf{Randomly select} a dimension $q \in \{1, \cdots, d\}$ according to distribution $D$\;
		\STATE \textbf{Randomly select} a split value $p$ between max and min
		values along dimension $q$ in $S$\;
		\STATE $S_l \leftarrow $ filter$(S, q < p)$\;
		\STATE $S_r \leftarrow $ filter$(S, q \ge p)$\;
		\RETURN inNode(Left $\leftarrow$ DiFF-Tree$(S_l , h + 1, h_{max})$,\\
		\hspace{10mm} Right $\leftarrow$ DiFF-Tree$(S_r , h + 1, h_{max})$,\\
		\hspace{10mm} splitAtt $\leftarrow q$,\\
		\hspace{10mm} splitVal $\leftarrow p$);
		\ENDIF		 
	\end{algorithmic}
	\caption{Function DiFF-Tree$(S, h, h_{max})$}
	\label{Algo:DiFF-Tree}
\end{algorithm}

\begin{algorithm}[]
	\begin{algorithmic}[1]
		\REQUIRE{$S \subset X_n$}\\
		\ENSURE{D, a probability distribution over the feature space dimensions $\{1, \cdots, d\}$}
		\IF{$|S| \le 10$}
		\RETURN $U[1/d]_{i \in \{1, \cdots d\}}$
		\ELSE
		\STATE $D \leftarrow $ $max([1-EE_i]_{i \in \{1, \cdots d\}},.2)$ (C.f.  Eq. \ref{eq:EE} for $EE_i$)\;
		\RETURN $D/\sum_{i=1}^{d}(D_i)$\;
		\ENDIF		
	\end{algorithmic}
	\caption{Function get\_qDistribution($S$): $EE_i$ is the empirical normalized entropy of dimension $i$ evaluated using an histogram with $\#bins=10$. $U$ stands for the uniform distribution.}
	\label{Algo:getDistrib}
\end{algorithm}

The partitioning algorithm used in DiFF-FR differs from that used by IF, in the way cutting dimensions are selected. It is somehow similar to the attribute selection used in an improvement of IF described in \cite{Liao2018}. The selection is obtained through the use of an empirical probability distribution $D$. Its justification is based on the following remark.

Dimensions with very high entropy can be assimilated to noise and therefore structurally less exploitable to partition an instance set. Hence, in DiFF-FR, we favor dimensions associated to low to medium entropy. To that end, we estimate empirically on the subset $S_n$ associated to the node to be split, the entropy $H_q$ of each dimension $q$. After applying the normalizing function $(1-H_q/log_2(\#bins))$, where $\#bins$ is the number of bins in the histograms, we obtain the probability of selecting a dimension, as depicted in Algorithm \ref{Algo:getDistrib}. For all our experimentation, to get a reasonable estimation, we fixed the number of bins to $10\%$ of the size of the data sample affected to a node with a maximum set to 100 bins and a minimum set to 5 bins. 

Basically, for each dimension an histogram is evaluated based on the instances in $S$. The empirical normalized entropy given in Eq. \ref{eq:EE} is then computed on the bins of this histogram, and finally normalized by the maximum entropy ($log_2(\#bins)$).
\begin{eqnarray}
\label{eq:EE}
\forall i \in \{1,\cdots d\} \text{, } \nonumber\\
EE_i=\frac{-1}{log_2(\#bins)}\sum_{k=1}^{\#bins}b_k/|S|\cdot log_2(b_k/|S|)
\end{eqnarray}

\subsubsection{Constructing the anomaly scores}

the anomaly score for DiFF-RF is constructed from the analysis of search results in the set of DiFF trees. Two cases are considered, depending on whether one is dealing with a point-wise anomaly or collective anomalies. We consider collective anomaly testing when a group of instances containing normal and abnormal samples in any proportion is tested simultaneously, as a whole. This may include sequential anomaly detection.\\

The point-wise score that we propose differs is similar to the local outlier score that is used as a post-processing step in \cite{Cheng19}. However, the local outlier factor requires a near neighbor search,  which is much costly when large datasets are involved. Following earlier attempts presented in \cite{Marteau2017} we implement a much more efficient distance based approach to a single centroid. Note here that a local clustering could provide several centroids at a cost that is linear with the number of clusters.   
\noindent \textbf{Score for point-wise anomaly detection}: for a given tree $T$, and a given point-wise data $x$ falling in a leaf $e$ of $T$ associated to subset of instances $S$, the anomaly score is defined through the fitting of a simple Gaussian model that results in the weighted distance $\delta(M_S, \sigma_S, x)$:
\begin{equation}
\delta(M_S, \sigma_{S}, x)=\frac{1}{d}\sum_{i=1}^{d}\Big(\frac{x(i)-M_S(i)}{\sigma_{S}}\Big)^2
\label{eq:weightedDistance}
\end{equation}
\noindent where $M_S$ and $\sigma_{S}$ are respectively the centroid and standard deviation of the training instances attached to the leaf in which $x$ falls. For tree $T$, the anomaly score is evaluated as 
\begin{equation}
\delta_T(x)=2^{-\alpha \cdot \delta(M_S, \sigma_{S}, x)}
\label{eq:delta_T}
\end{equation}

The point-wise anomaly score, \textit{pwas}, is then defined as:
\begin{equation}
pwas(x) = - \mathbb{E}\big(\delta_T(x)\big)
\label{eq:pws}
\end{equation}
\noindent where $\mathbb{E}$ is the mathematical expectation taken over the collection of
DiFF trees in the forest, and $\alpha$ is the single extra hyper-parameter used to scale the distance calculations.

Hence, when the expectation of the distances between $x$ and the leaf centroids tends toward $0$, the anomaly score $pwas(x)$ takes its minimal value, $-1$, while, when the expectation of these distances tends toward infinity, the score $pwas(x)$ takes its maximal value, $0$.

Algorithm \ref{Algo:Di_RFScoring} presents the recursive evaluation of $\delta_T(x)$ given $x$, a DiFF tree, $T$.\\

\begin{algorithm}[]
	\begin{algorithmic}[1]
		\REQUIRE{$x$ an instance, $T$ a DiFF tree}
		\ENSURE{$\delta_T(x,T,0)$, the point-wise anomaly score for $x$ provided by $T$}
		\IF{$T$ is a leaf node associated to substet of instances $S$}
		\RETURN $2^{-\alpha \cdot \delta(M_S, \sigma_{S}, x)}$\\
		\ENDIF	
		\STATE $a \leftarrow T.splitAtt$;\\
		\IF{$x[a] < T.splitValue$}
		\RETURN $\delta_T(x, T.left)$;
		\ELSE
		\RETURN $\delta_T(x, T.right)$;\\
		\ENDIF	
	\end{algorithmic}
	\caption{Function $\delta_T(x)$}
	\label{Algo:Di_RFScoring}
\end{algorithm}

\noindent \textbf{Score for collective anomaly detection}:
as defined in \cite{Chandola:2009}, collective anomalies occur when a subset of related data instances is abnormal relatively to the training data set. Notice that collective anomalies can be composed with point-wise normal instances. It is then the abnormal co-occurrences of these instances that characterize the collective anomaly.  
DiFF-RF considers the visiting frequencies in the leaf nodes as the basic element for constructing a collective anomaly score. When constructing the trees, the estimated frequency of visit in a leaf node $e$ is evaluated as the ratio $f_{n} = |S|/\psi$ between the number of instances attached to the leaf $e$, $|S|$, and the total number of training instances used to build the tree, namely $\psi$. This is depicted at line $2$ of Algorithm \ref{Algo:DiFF-Tree}.\\

At test time, when a subset $X \subset R^d$ of instances, potentially containing some collective anomalies, a new frequency of visit is evaluated at each leaf $e$ as $f_{X}=|S_X|/|X|$, where  $S_X$ is the subset of elements of $X$ that fall in leaf $e$.

For each leaf $e$ of each tree $T$ in the forest, we can thus evaluate the relative train/test visit frequencies at leaf levels as the ratio $\nu_{T}(X) = f_{n}/f_{X}$. For tree $T$, the collective anomaly for instance $x \in X$ is calculated as the aggregation of the distance score and the relative frequency score: $c_T(x, X) = \delta_T(x)\cdot \nu_{T}(X)$

Finally, the collective anomaly score given the context $X$ is:

\begin{equation}
cas(x,X) = -\mathbb{E}\big(c_T(x,X)\big)
\label{eq:cas}
\end{equation}

\noindent where $x \in X$ and $\mathbb{E}$ is the mathematical expectation taken over the trees $T$ in the forest.

Hence, when the ratio of visit frequencies tends towards $0$, i.e. when the leaves are much more visited during test time, then the score $cas(x,X)$ tends to its maximum, $0$, which will characterized  the presence of collective anomalies. 

Algorithm \ref{Algo:DiFF_RFScoring} presents the recursive evaluation of $c_T(x,X) = \delta_T(x)\cdot \nu_T(x,X)$ given $x$, $X$, and a DiFF tree, $T$, the current path length, $h$ being initialized with $0$.\\

\begin{algorithm}[]
	\begin{algorithmic}[1]
		\REQUIRE{$X$ a subset of test instances, $T$ an DiFF tree}
		\ENSURE{$\nu_T(x, X, T, 0)$, the collective anomaly score for all $x \in X$}
		\IF{$T$ is a leaf node associated to substet of instances $S$}
		\RETURN $\forall x \in X$, $2^{-\alpha \cdot \delta(M_S, \sigma_{S}, x)} \cdot f_n/f_X$\\
		\ENDIF	
		\STATE $a \leftarrow T.splitAtt$;\\
		\IF{$x[a] < T.splitVal$}
		\RETURN $\nu_T(x, X, T.left)$;
		\ELSE
		\RETURN $\nu_T(x, X, T.right)$;\\
		\ENDIF	
	\end{algorithmic}
	\caption{Function $\nu_T(x, X)$}
	\label{Algo:DiFF_RFScoring}
\end{algorithm}

Of course, this type of scoring will not be effective if collective anomalies are hidden within a large amount of normal data. However, in the case of streaming data, the use of a sliding window would make it possible to detect some significant variations in these relative frequencies of occurrence. Such variations are cues of the presence of collective anomalies.

\subsection{`Blind spots' in IF do not exist in DiFF-RF}
The assumption behind the IF algorithm is that anomalies will be associated to short paths in the partitioning trees, leading to a high anomaly score, while 'normal' data will be associated to longer paths, leading to a low anomaly score. Unfortunately, if this is true for normally distributed data for instance, this is not true in general. In particular, this assumption is not verified for data distributed in a concave set such as a torus or a set with a 'horse shoe' shape. This `blind spot' effect is greatly reduced in the DiFF-RF because of the distance criteria to the centroid evaluated at the leaf nodes. To demonstrate this, we develop the following experiment based on synthetic data.

\subsubsection{Synthetic experiment} \label{donnut}

\begin{figure*}[h!]
	\centering
	\subfigure[]{
		\includegraphics[width=0.2\textwidth]{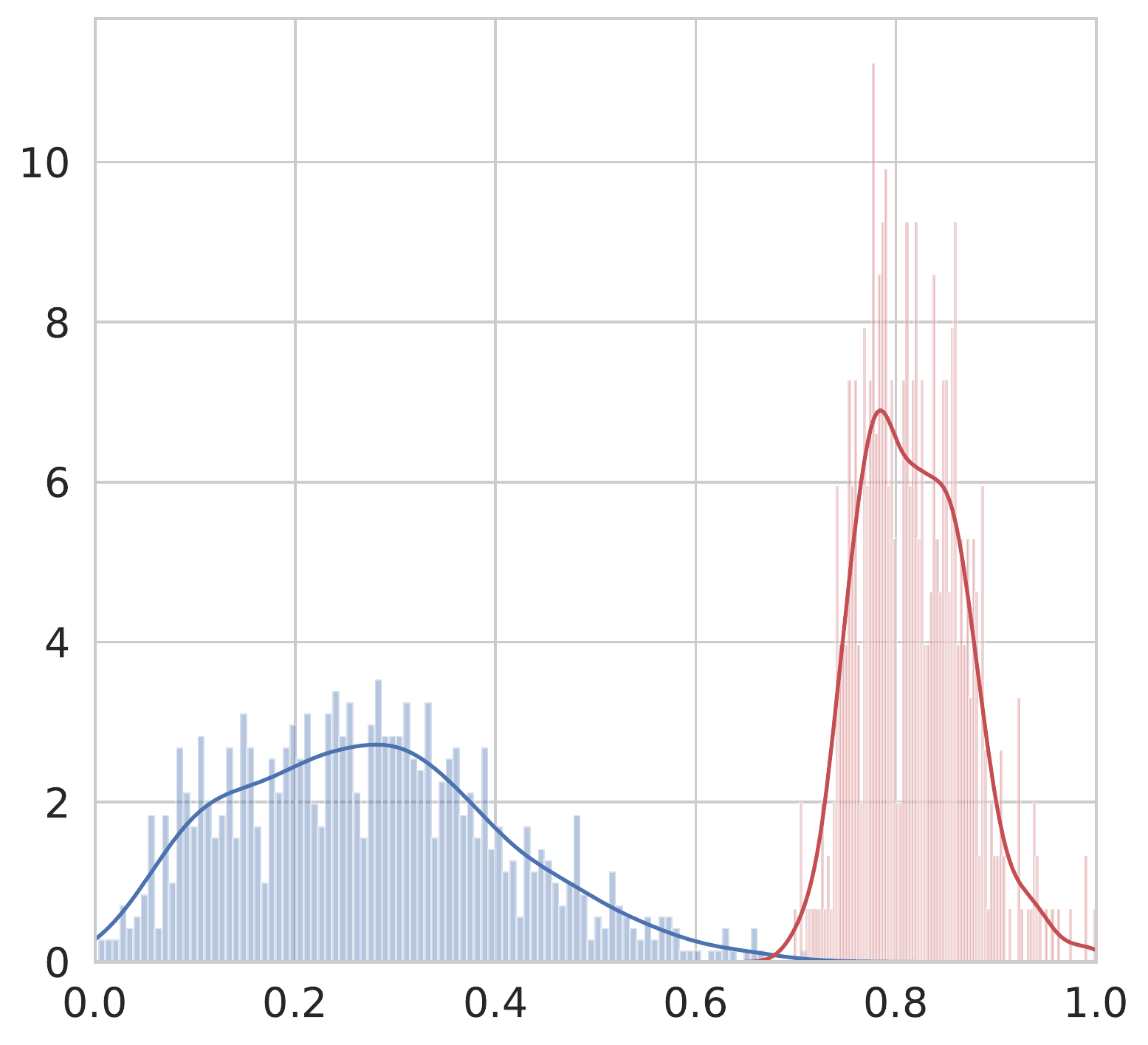} 
	}
	~
	\subfigure[]{
		\includegraphics[width=0.2\textwidth]{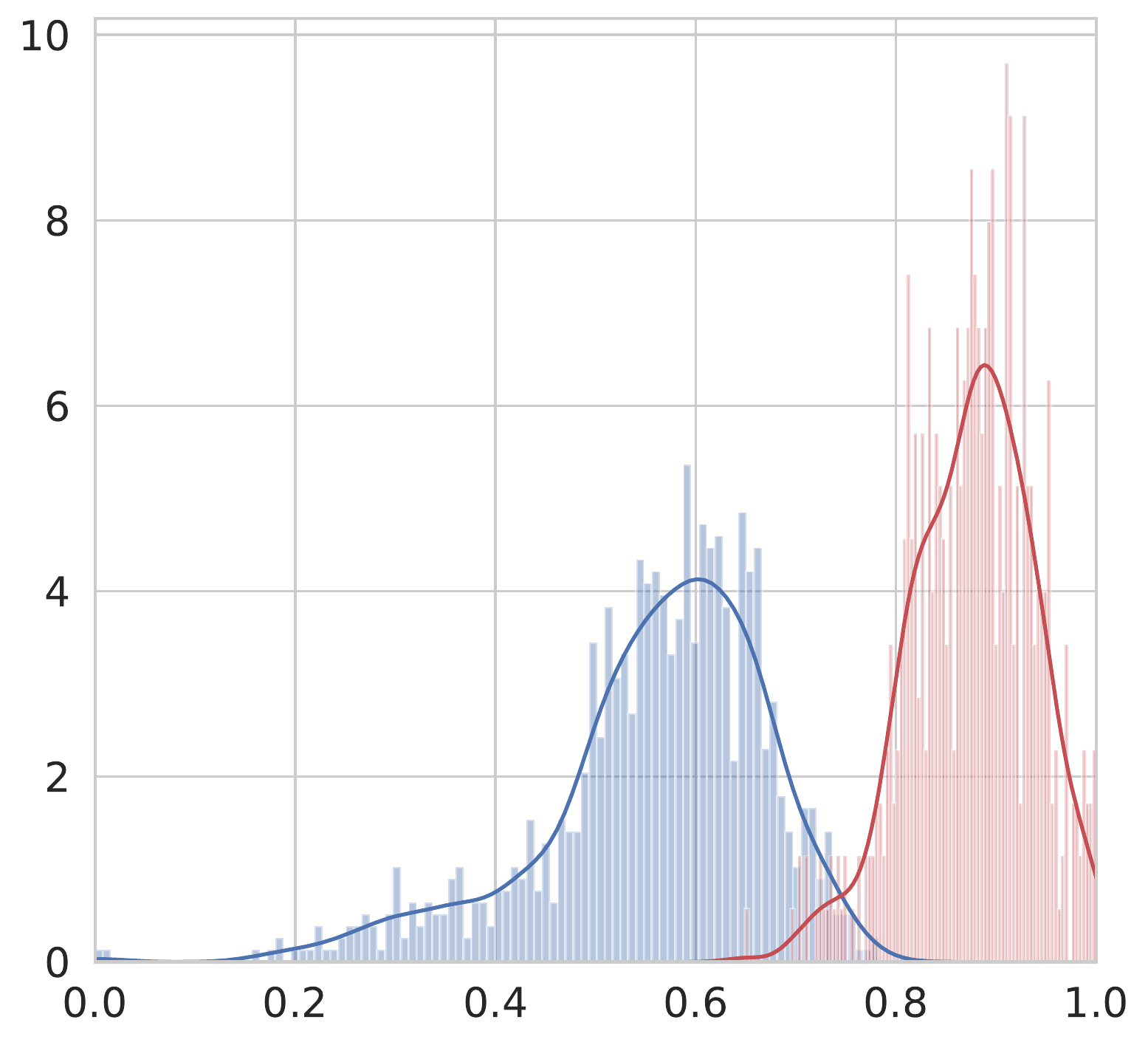} 
	}
	~
	\subfigure[]{
		\includegraphics[width=0.2\textwidth]{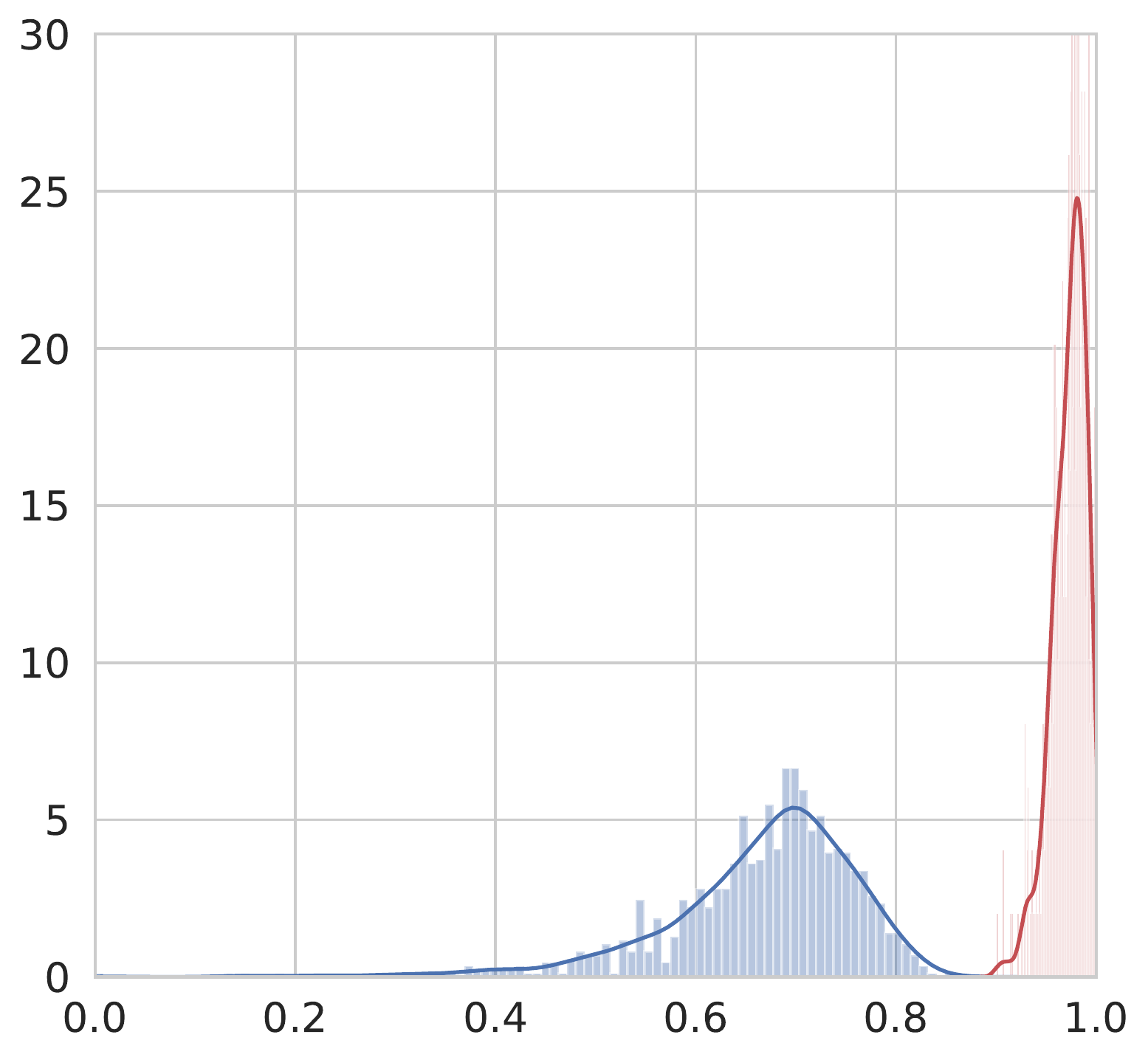}
	}\\
	\subfigure[]{
		\includegraphics[width=0.2\textwidth]{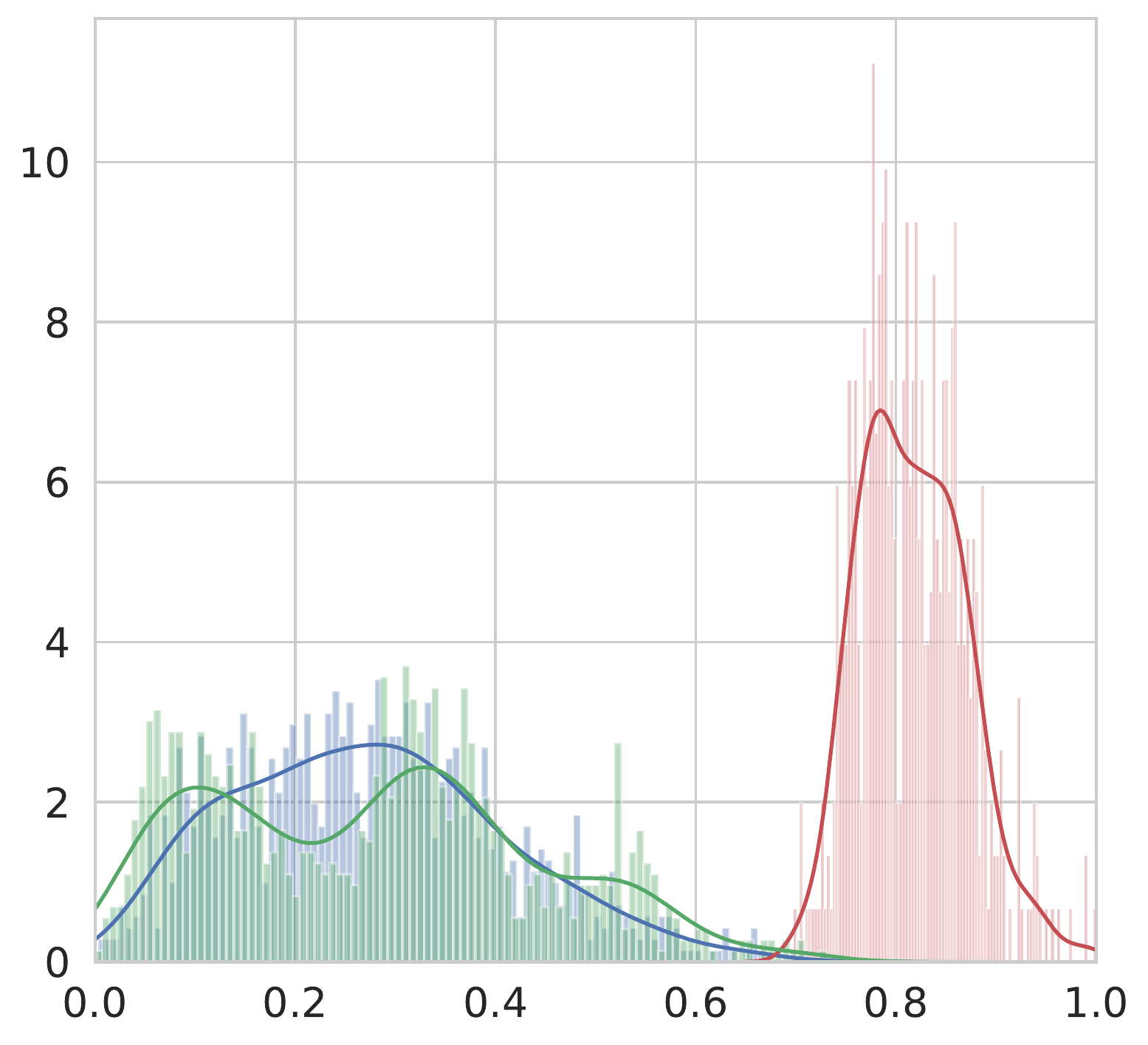}
	}
	~
	\subfigure[]{
		\includegraphics[width=0.2\textwidth]{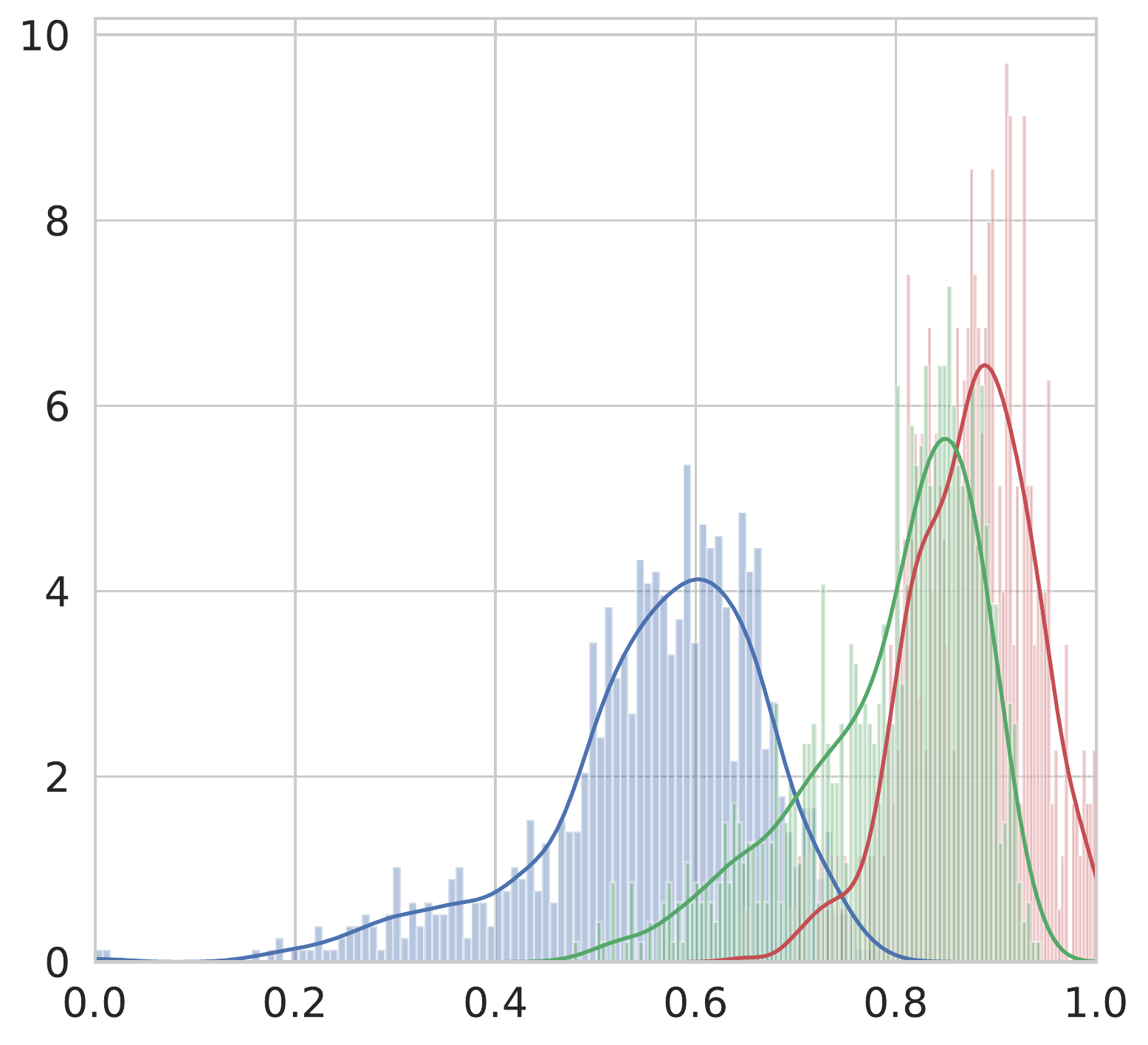} 
	}
	~
	\subfigure[]{
		\includegraphics[width=0.2\textwidth]{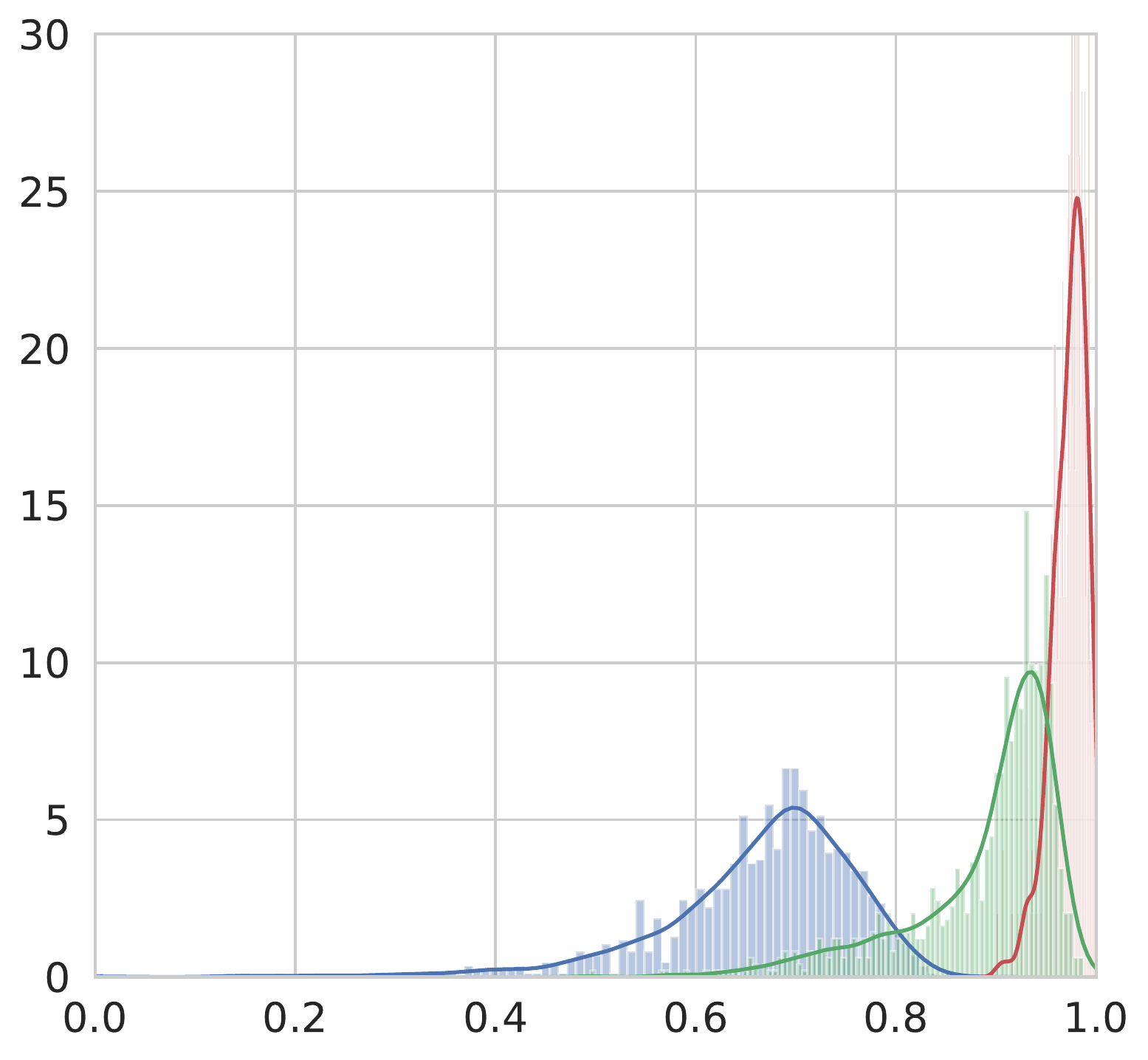}
	}
	\caption{Distributions of the anomaly scores for the 'normal' data, in blue, and for the red anomalies (top), and for the green anomalies (bottom). The IF distributions are given in (a),(d) sub-figures, DiFF\_RF point-wise anomaly distributions are given in (b), (e) sub-figures and the DiFF\_RF collective anomaly distributions are given in (c), (f) sub-figures.}
	\label{fig:distrib-DiFF-RF}
\end{figure*}
\begin{figure}[h!]
	\centering
	\includegraphics[scale=0.2]{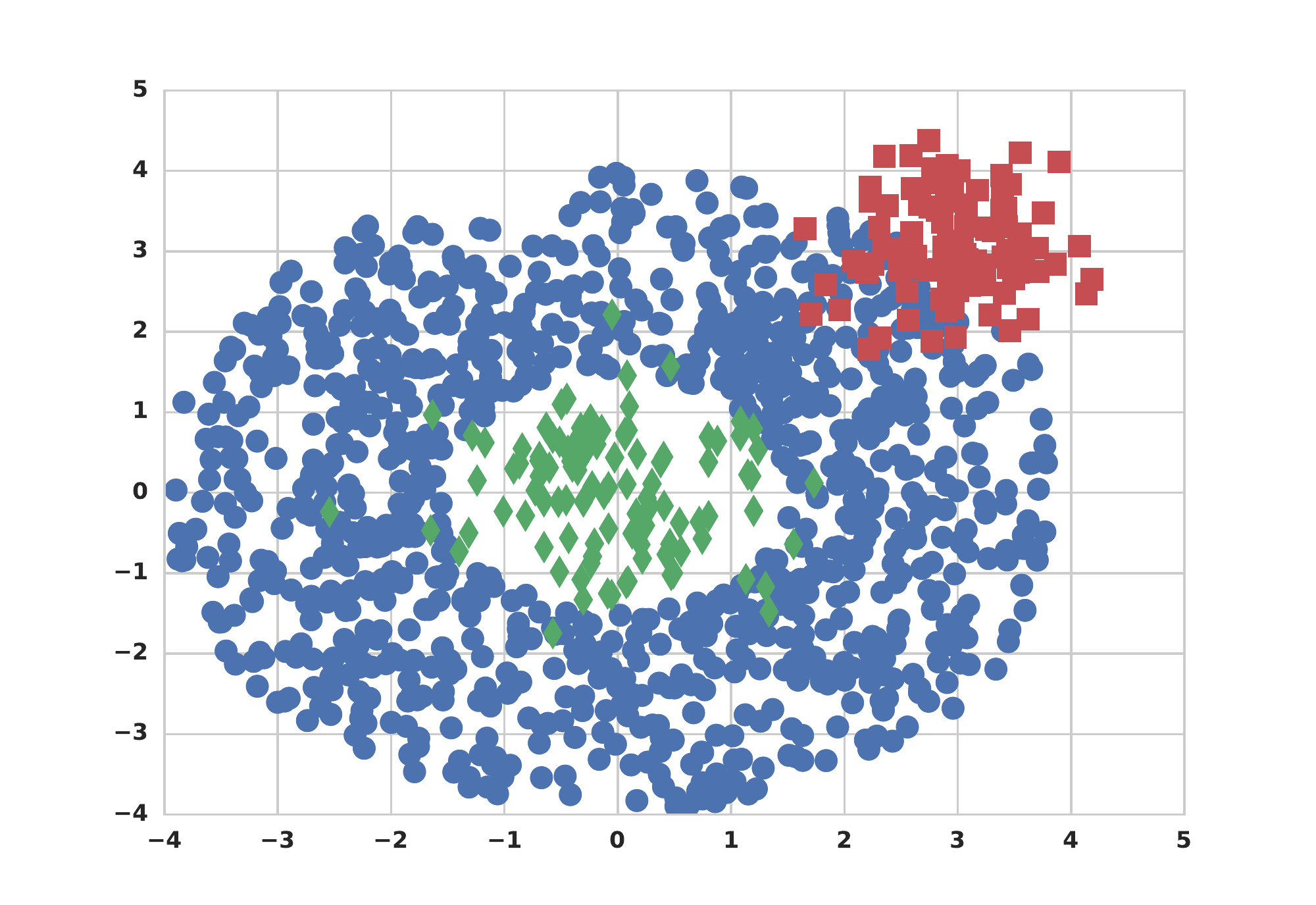} 
	\caption{2D-torus 'normal' data set in blue round dots, with a cluster of anomaly data in red square dots at the top right side of the torus, with an additional cluster of anomaly data in green diamond dots at the center of the torus.}
	\label{fig:clusters}
\end{figure}


in this setting, 'normal' data belongs to a 2D-torus centered in $(0,0)$ and delimited by two concentric circles whose radius are respectively $1.5$ and $4$. A training ($X_n$) and a testing ($X_{nt}$) sets of normal data are uniformly drawn from this 2D-torus, each one containing $n=1000$ instances, as depicted in Fig. \ref{fig:clusters}. 

A first 'anomaly' set ($X_{r}$) is drawn from a Normal distribution with mean $(3., 3.)$ and covariance $((.25, 0), (0, .25))$, as depicted in red square dots at the top right side of Fig. \ref{fig:clusters}. These anomalies intersect the 2D-torus at its top right side.

A second 'anomaly' set ($X_{g}$) is drawn from a Normal distribution with mean $(0., 0.)$ and covariance $((.5, 0), (0, .5))$, as depicted in green diamond dots located in the center of the 2D-torus in Fig. \ref{fig:clusters}. 

Then we build the IF and the DiFF-RF (with a sample size $\psi=512$ and $t=128$ trees) from the 'normal' dataset $X_{n}$ and evaluate the distributions of the anomaly scores obtained for the 'normal' 'blue' test dataset $X_{nt}$, the 'red' anomalies $X_{r}$ and the 'green' anomalies $X_{g}$. In Fig.\ref{fig:distrib-DiFF-RF}, the left column presents the IF algorithm the 'normal' v.s. 'red' anomalies (a)  distributions, and with the addition of the green anomaly distribution (c). \\


At this point, we clearly show that for IF the 'green' anomaly distribution is in large intersection with the 'normal' data distribution, which is not the case for the 'red' anomaly distribution. Hence anomalies located at the center of the torus are likely to be much more mis-detected by the IF algorithm than anomalies located at the periphery of the torus.\\

Similarly,  2nd to 4th columns of Fig.\ref{fig:distrib-DiFF-RF} present the DiFF-RF algorithm scores for the 'normal' v.s. 'red' anomalies (top)  distributions, and with the addition of the green anomaly distribution (bottom). The 2nd column corresponds to the point-wise anomaly scores, the 3rd column corresponds to the expectation of the ratio of visiting frequencies at the leaf nodes ($\nu_{T}(X)$) and the right column corresponds to the collective anomaly scores.\\

We can see that, thanks to the distance-based measure, the point-wise anomaly score is able to discriminate the green anomaly as well as the red ones, although not perfectly since the distributions still overlap. The frequency of visits seems to reasonably discriminate the red anomaly but also suffers from a `blind spot' effect. However, the aggregation of the distance score with the frequency score is particularly discriminating for both types of anomalies (red and green). This shows a great complementarity of these two scores in the context of collective anomaly detection.

\begin{figure*}[h!]
	\centering
	\subfigure[]{
		\includegraphics[scale=0.24]{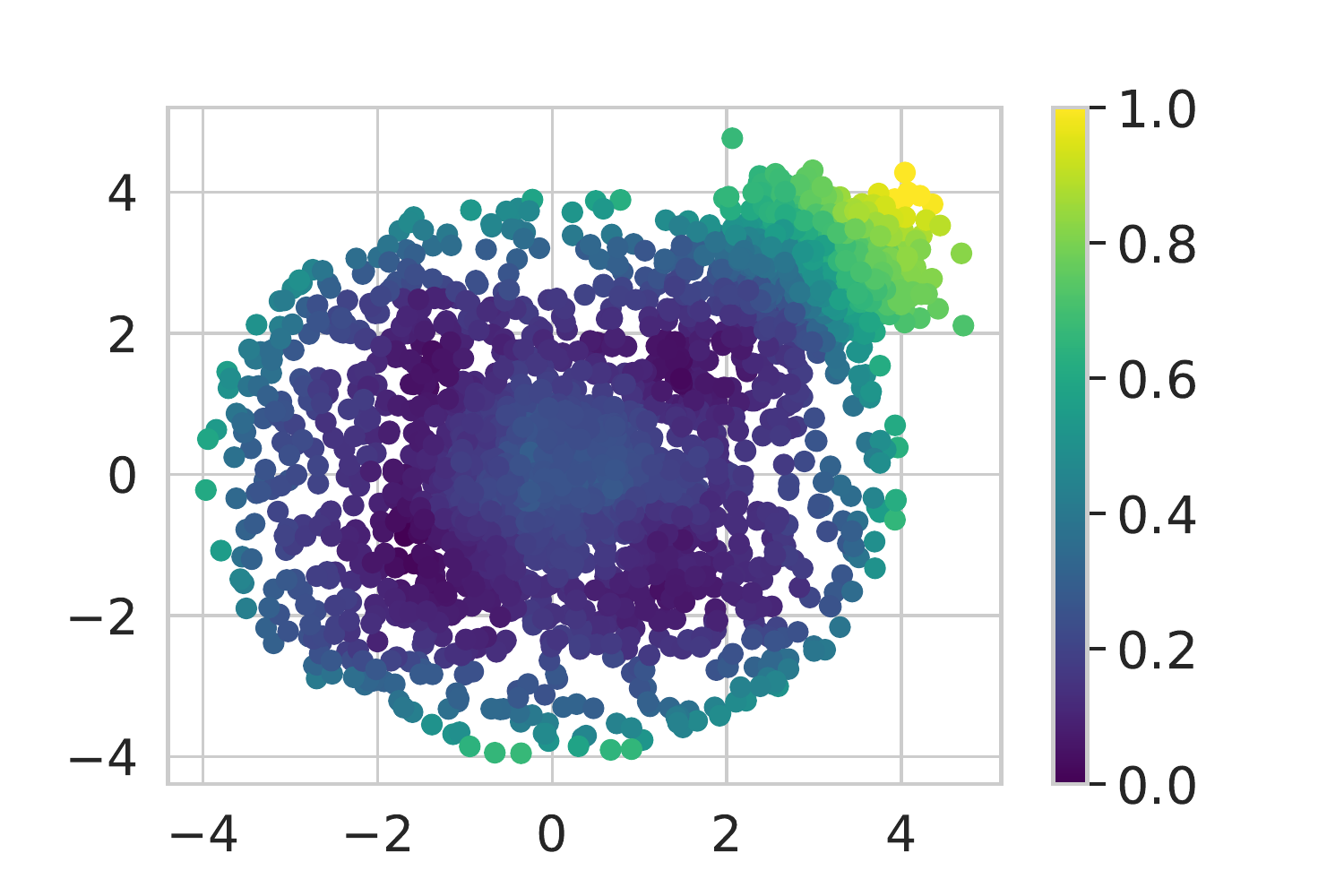} 
	}
	~
	\subfigure[]{
		\includegraphics[scale=0.24]{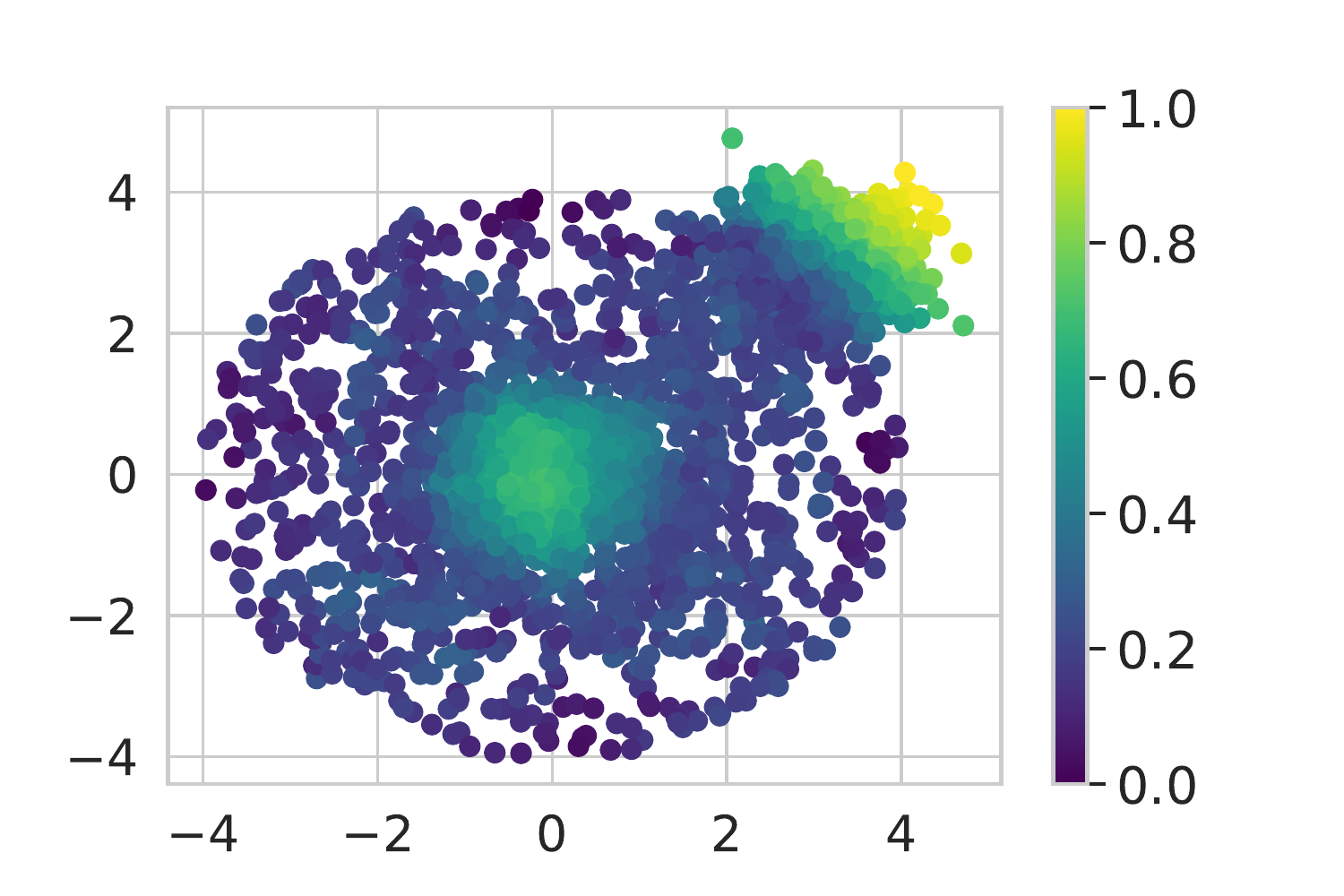} 
	}
	~
	\subfigure[]{
		\includegraphics[scale=0.24]{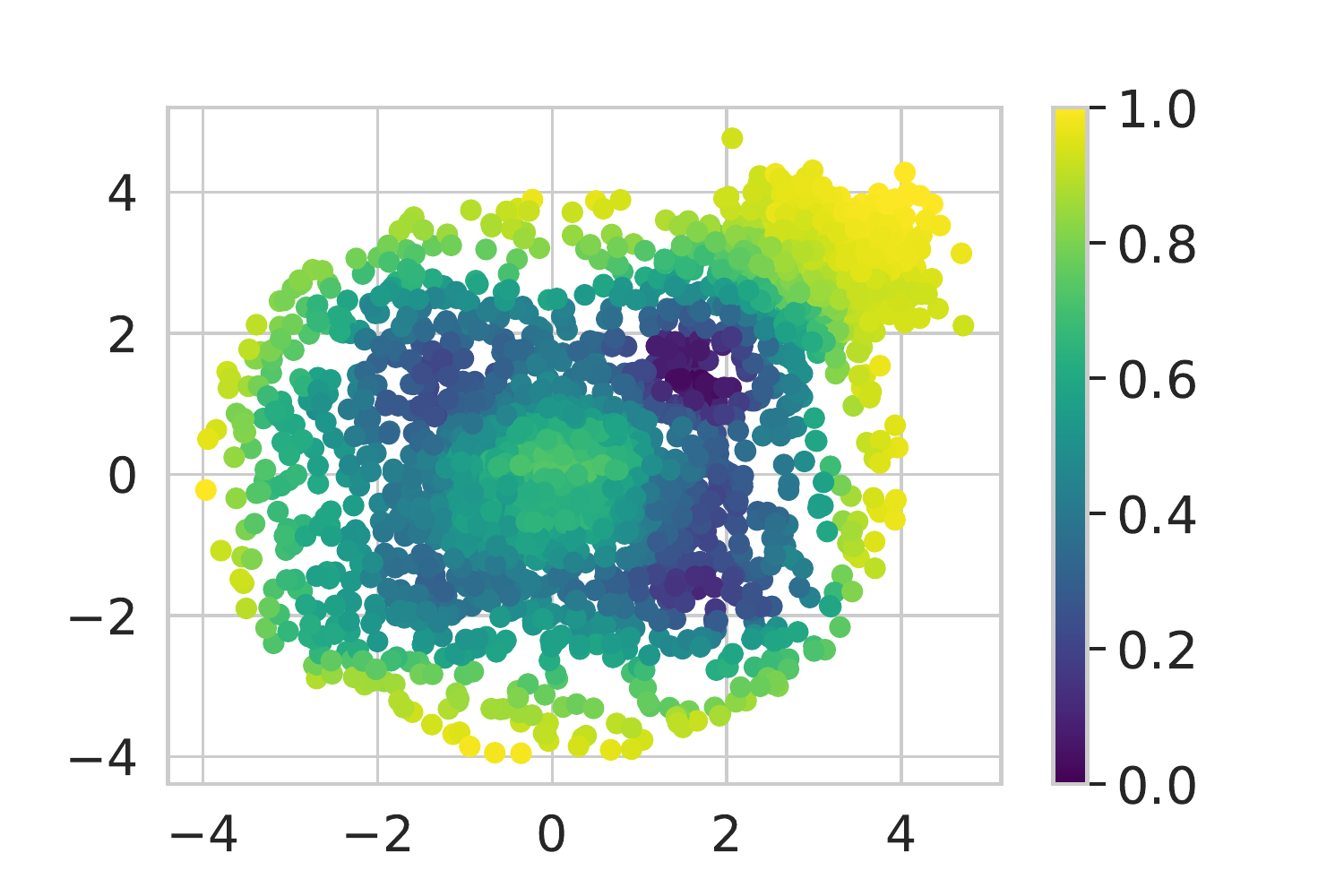} 
	}
	~
	\subfigure[]{
		\includegraphics[scale=0.24]{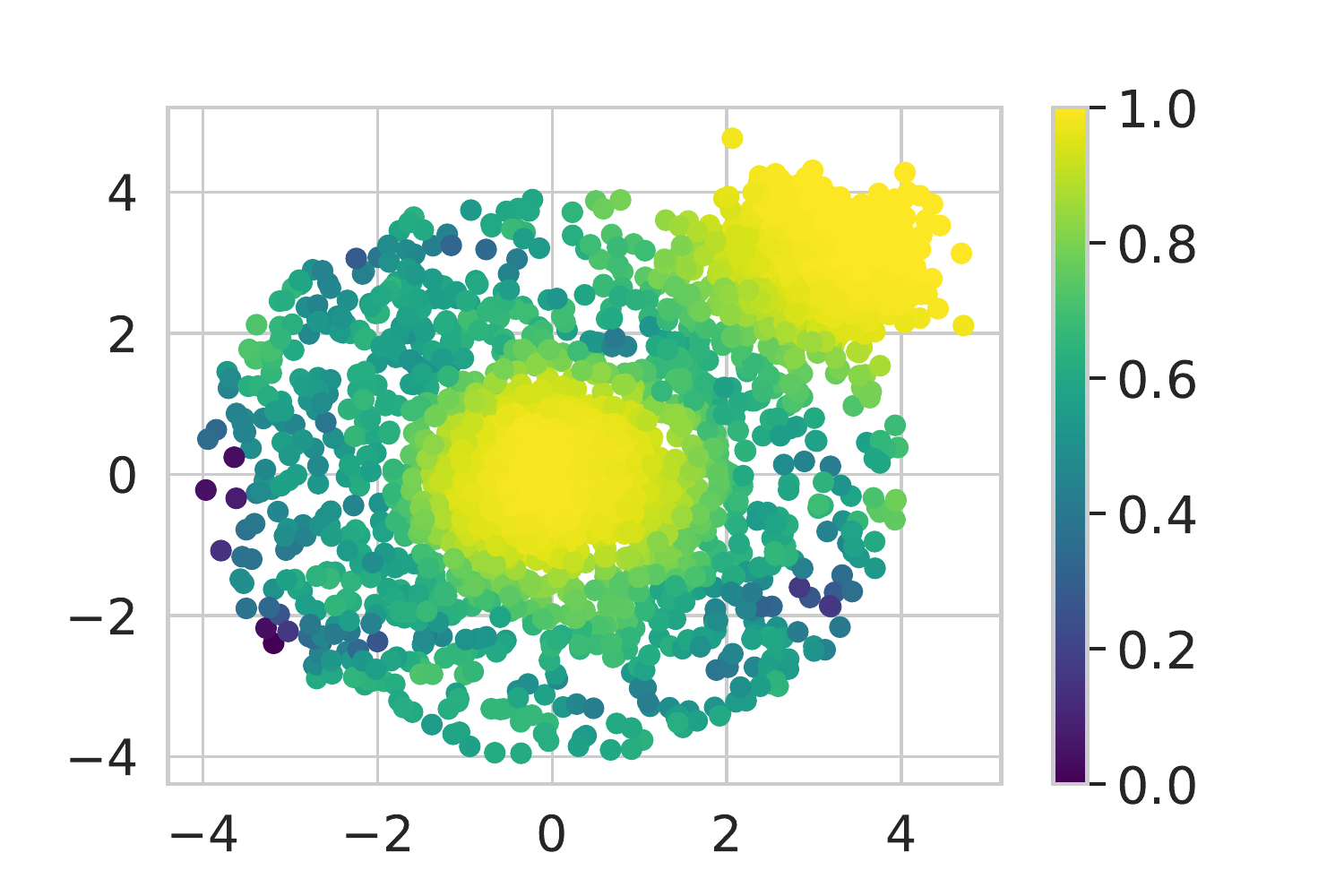} 
	}
	
	\caption{2D-torus heat map corresponding to the IF (a), to the point-wise anomaly score of the DiFF-RF (b), to the expectation of the ratio of visiting frequencies in the DiFF-RF (c), to the collective anomaly score of the DiFF-RF (d).}
	\label{fig:heatmaps}
\end{figure*}

Fig. \ref{fig:heatmaps} provides the heat maps evaluated for the IF score (\ref{fig:heatmaps}-a), and the DiFF-RF scores (point-wise anomaly score (\ref{fig:heatmaps}-b), expectation of the ratio of visiting frequencies in the DiFF-RF (\ref{fig:heatmaps}-c) and to the DiFF-RF collective anomaly aggregated score (d)). We clearly visualize the `blind spot' effect for the IF (\ref{fig:heatmaps}-a), and not for the point-wise anomaly score of the DiFF-RF (\ref{fig:heatmaps}-b). The heat map corresponding to the ratio of visiting frequencies is quite interesting: the hottest points (yellow/light) are located at the limits of the torus and on the anomaly clusters whose instances are likely to fall in leaves which are associated to training instances precisely located at the limits of the torus. The product of these two complementary scores provides obviously a very discriminating score, able to separate neatly on this experiment the two types of anomalies from normal data, as shown in sub-figure (\ref{fig:heatmaps}-d).

\begin{figure}[h!]
	\centering
	\includegraphics[width=0.35\textwidth]{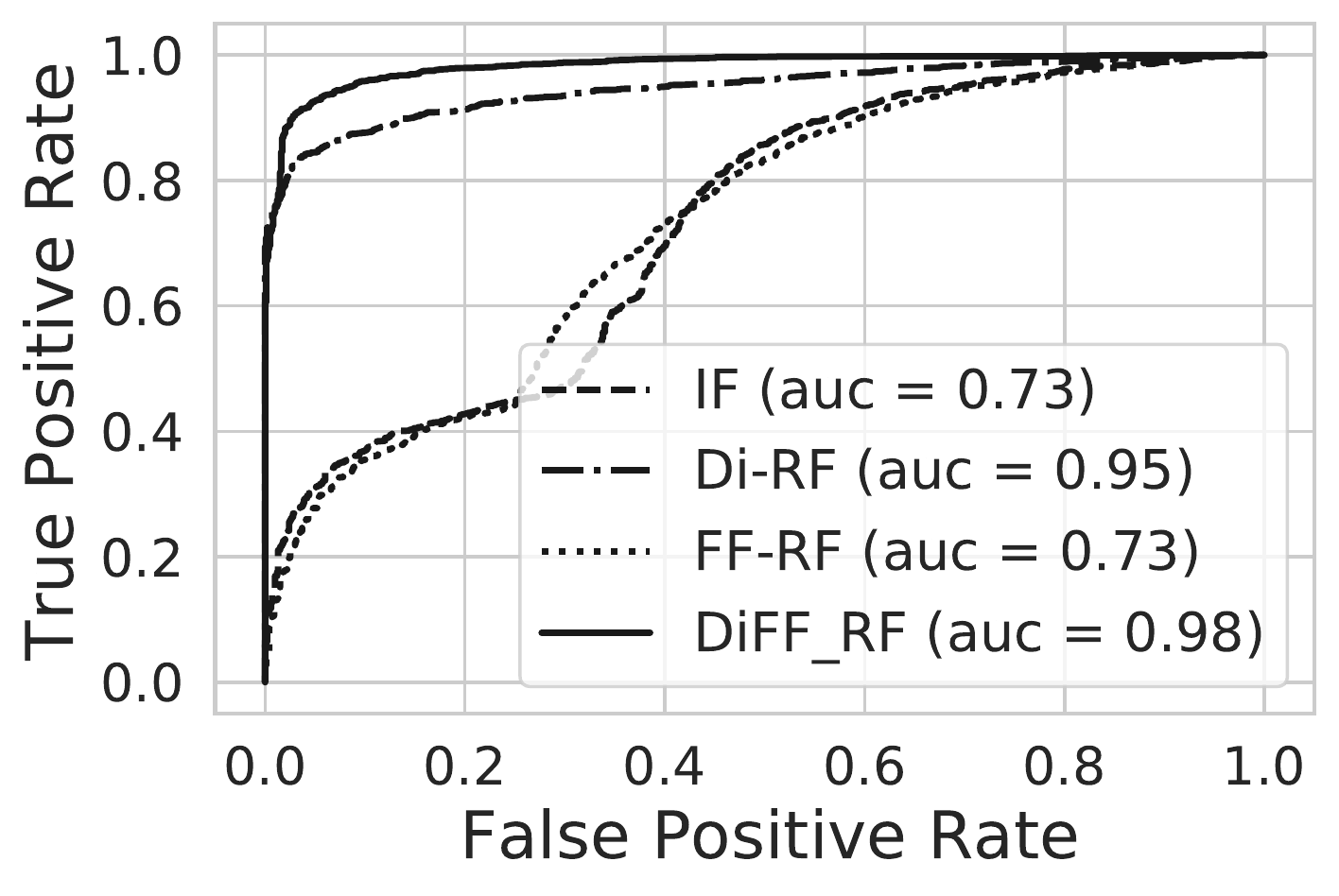} 
	\caption{ROC curves and AUC for IF (dashed line), DiFF-RF point-wise anomaly (Di-RF, dashdot line), DiFF-RF ratio of visiting frequency (FF-RF, dotted line), DiFF-RF collective anomaly (solid line): 'normal' test data against all anomalies (red and green). }
	\label{fig:roc1}
\end{figure}

To complete these results, Fig.\ref{fig:roc1} gives the Receiver Operating Curve (ROC) and Area Under the Curve (AUC) values for the tested detectors trained on 'normal' data only and tested on a disjoint set of normal data concatenated with the red and green anomalies subsets. For IF, due to the `blind spot' mis-detection, AUC is only $0.73$ while it reaches $.95$ for DiFF-RF point-wise detection and $.98$ for the DiFF-RF collective anomaly score.  For completeness, the AUC for the expected ratio of visiting frequency scoring is $.73$, indicating its discriminative complementarity with the distance-based score. These results are fully in accordance with our previous observations.\\


\subsubsection{Dependence to the sample size ($\psi$) in each tree and to the number of trees ($t$) in the forest} 
\begin{figure}[h!]
	\centering
	\includegraphics[width=0.45\textwidth]{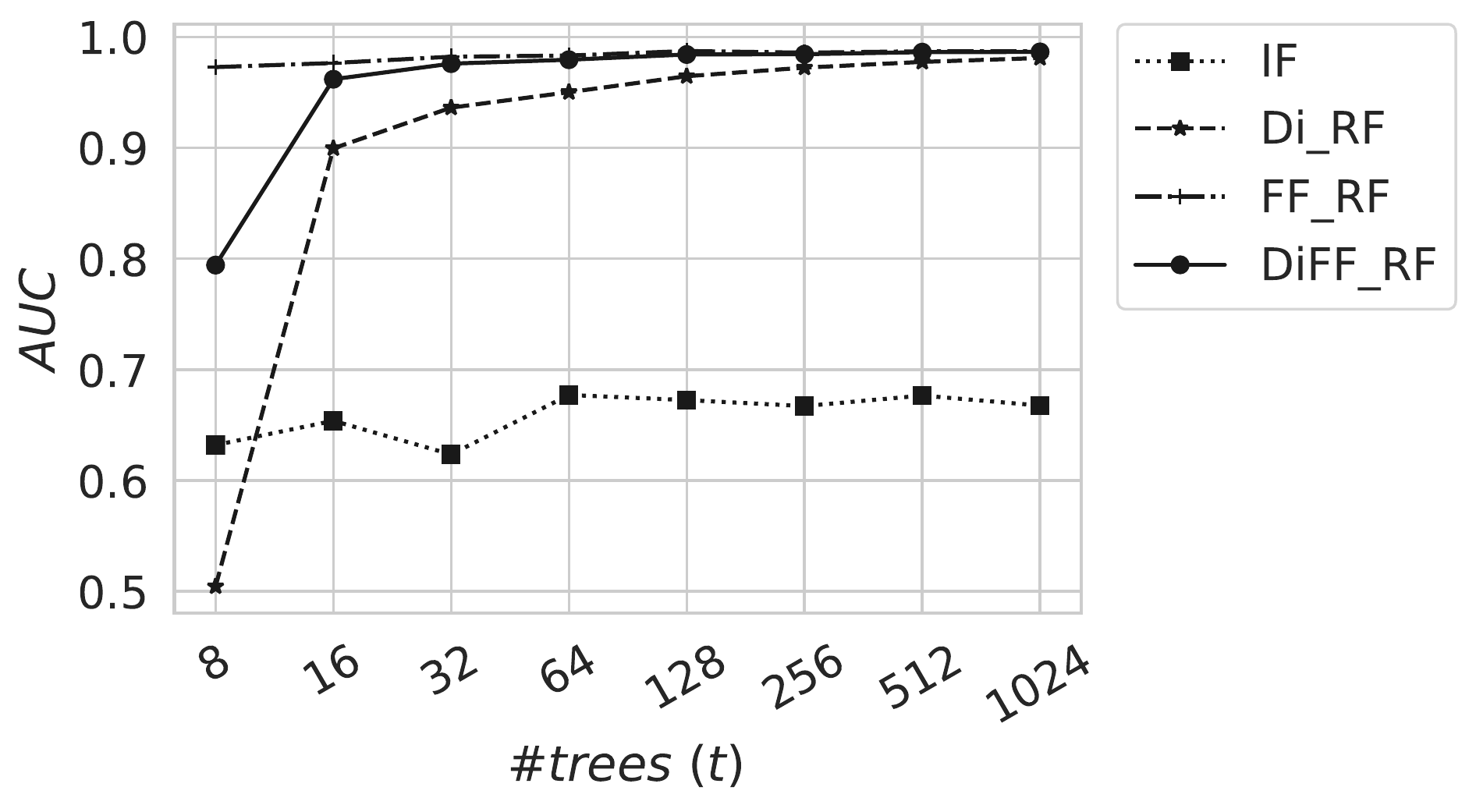}  
	\caption{AUC values when the number of trees varies while the sample size remains constant equal to 128 instances and $\alpha=10$.}
	\label{fig:t_meta_parameter}
\end{figure}

\begin{figure}[h!]
	\centering
	\includegraphics[width=0.45\textwidth]{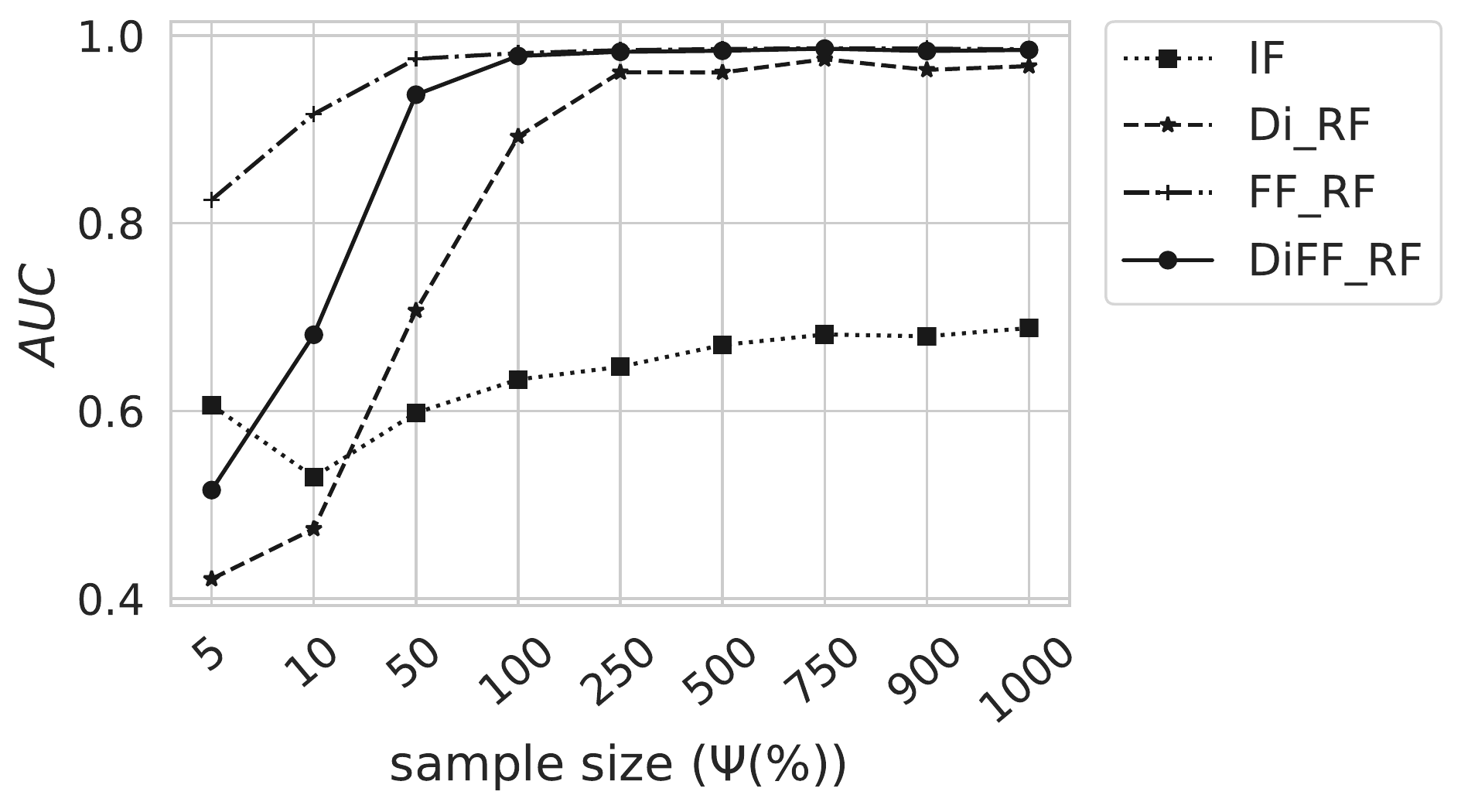}  
	\caption{AUC values when the sample size varies while the number of trees in the forest remains constant equal to 128 trees and $\alpha=10$.}
	\label{fig:psi_meta_parameter}
\end{figure}

the dependence of the AUC assessment measure to the hyper-parameter settings, namely the number of DiFF trees $t$, and the sample size $\psi$ assigned to each DiFF tree, are presented respectively in in Fig. \ref{fig:t_meta_parameter} and Fig. \ref{fig:psi_meta_parameter}.

For this experiment, a dataset size of $n=2000$ 'normal' samples is used to train the isolation forest, and $\alpha=10$ is maintained constant. One can see that the DiFF-RF in its point-wise detection configuration (Di-RF) reaches good and relatively stable AUC values with few trees (Fig. \ref{fig:t_meta_parameter}), from 128 to 1024 trees, and low sample sizes (Fig. \ref{fig:psi_meta_parameter}), from 250 to 1000 samples, which is quite advantageous in terms of memory space and response time. In its collective anomaly configuration (DiFF-RF), the performance are surprisingly high and almost constant whatever these two hyper-parameters are respectively above 32 trees and 100 instances.

\subsection{Setting up meta parameters $\alpha$}
\label{sec:meta_alpha}

\begin{figure}[h!]
	\centering
	\includegraphics[width=0.45\textwidth]{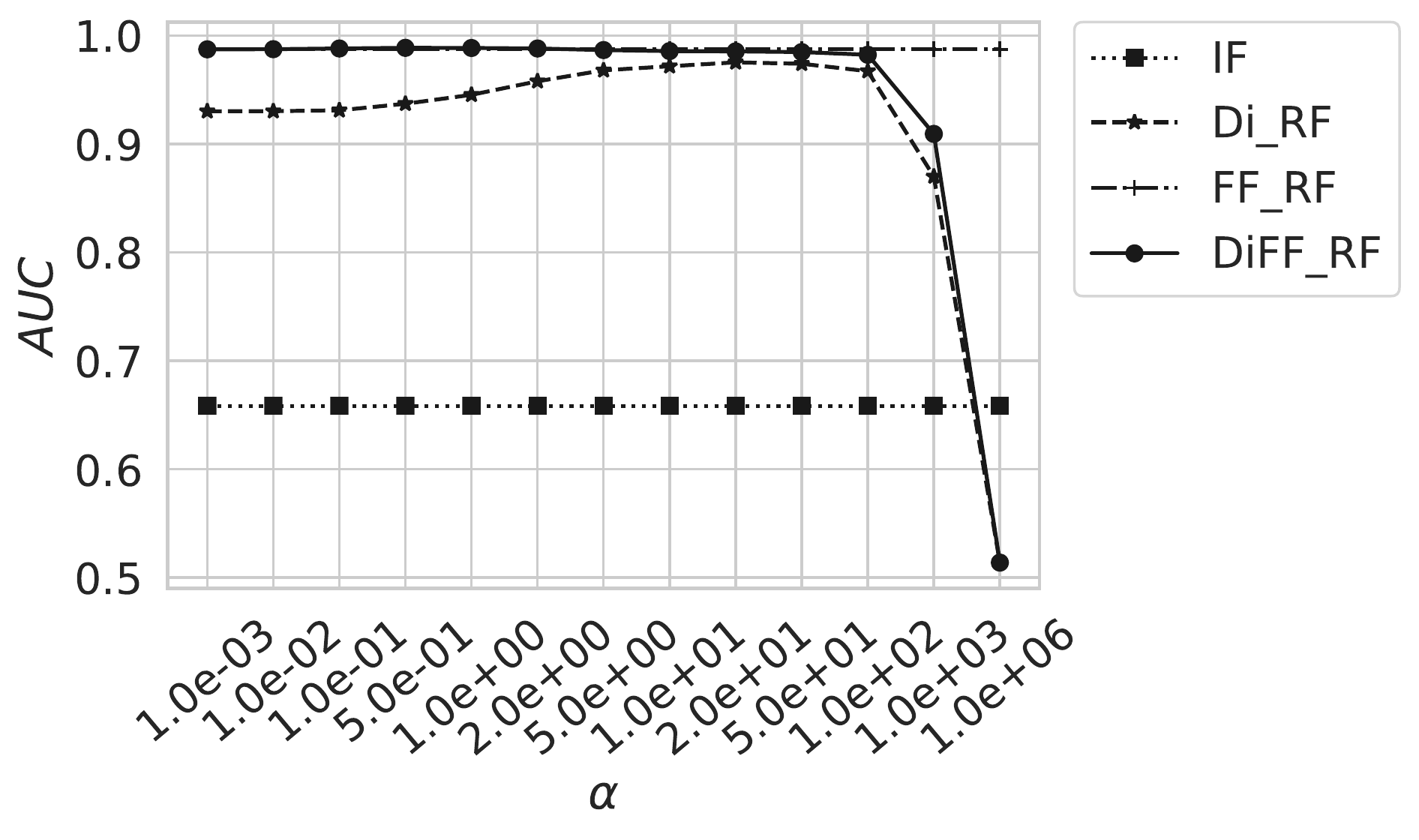}  
	\caption{AUC values when the $\alpha$ hyper-parameter value varies while the number of trees in the forest and the sample size remain constant equal to 128 trees and 256 instances respectively.}
	\label{fig:alpha_meta_parameter}
\end{figure}

Figures \ref{fig:alpha_meta_parameter} presents the AUC values as $\alpha$ varies in $[10^{-3};10^{3}]$. 
Basically, $\alpha$ is mainly used to ensure that the term $\delta_T$ (Eq. \ref{eq:delta_T}) remains computable under the experimental conditions encountered. The figure shows that, on this experiment, until $\alpha$ is not too high (below 100), $\delta_T$ is computable hence suitable. However, the figure shows that 'optimal' values may exists due to the non linearity of the exponential function. Hence this parameter may need some tuning to adapt to the application data (dimensionality of the problem and scale of the features). 

To tune and select the $\alpha$ parameter, we adopt a straightforward \textbf{semi-supervised} cross-selection procedure depicted in Algorithm \ref{Algo:AlphaTuning}. This procedure partitions training normal data into pairs of train/test sets, then evaluates the anomaly scores obtained on the train and test data and finally calculates a distance measure, $\delta_Q$, between the distributions of anomaly scores obtained, while considering only the highest scores, in order to focus on the boundary of "normality" as assumed by the method. $\delta_Q$ that is used at line 10 of Algorithm \ref{Algo:AlphaTuning} is defined as follows

\begin{equation}
\delta_Q(P_1,P_2)=\sum_{i=95}^{99} | |S_{q_i}|-(100-i) |
\label{eq:distribDist}
\end{equation}

where $|S_{q_i}|$ is the cardinal of the set of the instances of $P_1$ whose scores are above the score of the $i^{th}$ quantile of instances in $P_2$.\\

The $\alpha$ value that minimizes $\delta_Q$ is the one that is used during the testing phase.

\begin{figure}[h!]
	\centering
	\includegraphics[width=0.38\textwidth]{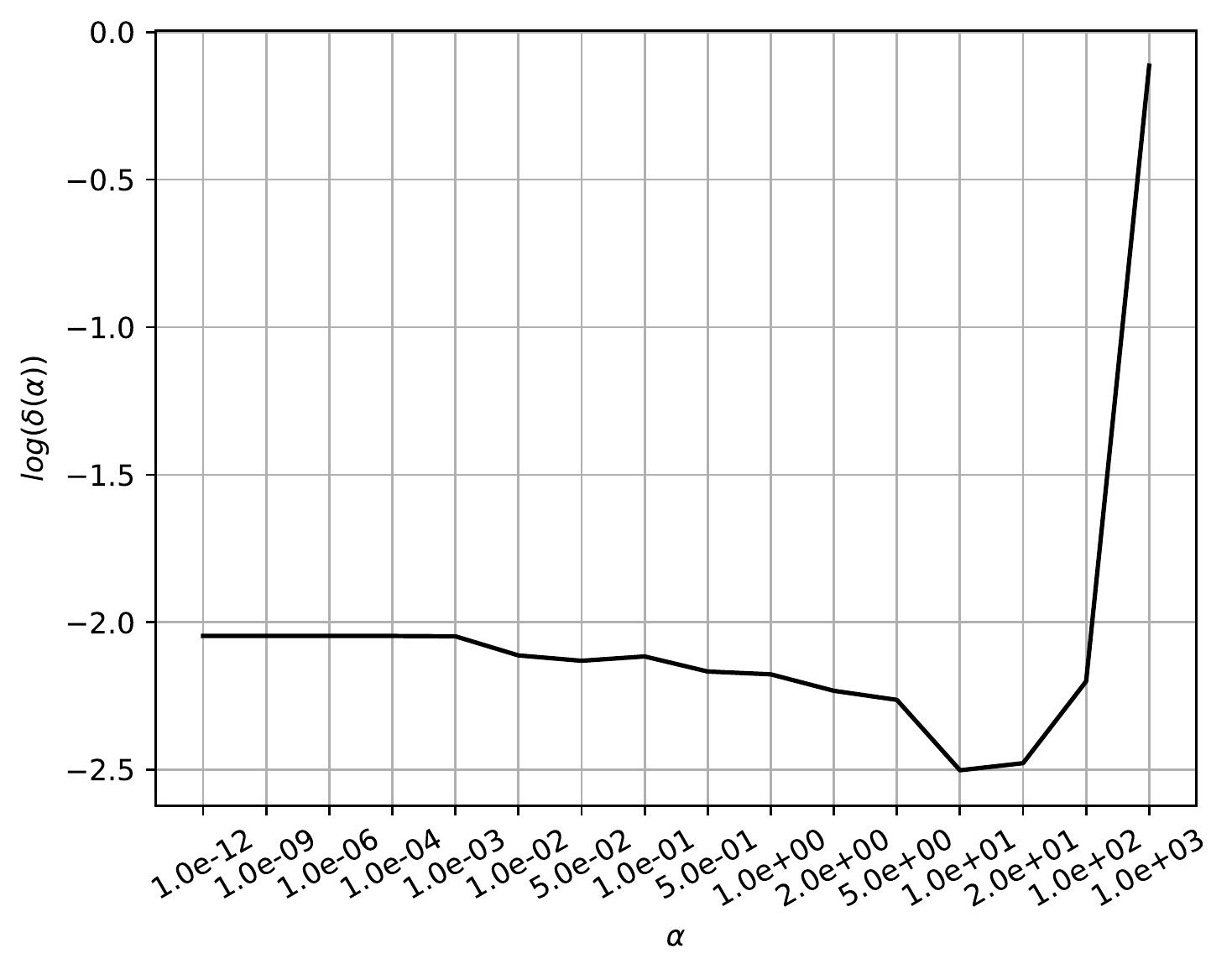}  
	\caption{$\delta_Q$ as a function of $\alpha$. This curve has been obtained after 12 iterations on the Donut synthetic data set using $t=256$ and $\psi = 256$.}
	\label{fig:deltaQ}
\end{figure}

This procedure applied on the previous synthetic data selects  $\alpha=10$, as shown in Fig. \ref{fig:deltaQ}.

\begin{algorithm}[]
	\begin{algorithmic}[1]
		\REQUIRE{$X$ a subset of 'normal' instances, $t$ the number of trees, $\psi$ the sample size, $\#iter$ the number of iterations.}
		\ENSURE{$\alpha$, the hyper-parameter defined in Eq. \ref{eq:delta_T}}
		\STATE $S_\alpha \leftarrow$ $[$1e-12, 1e-9, 1e-6, 1e-4,1e-3, 1e-2, .05, .1, .5, 1, 2, 5, 10, 100$]$;\\
		\STATE $\forall \alpha \in S_\alpha$, $R(\alpha)=0$ ;\\
		\FOR{$k=0$ to $\#iter$}
		\STATE $X \leftarrow$ $shuffle(X)$;\\
		$P \leftarrow$ $Partition(X,\psi)$ // Partition $X$ in elements of size almost equal to $\psi$;
		\FOR{$i=0$ to $|P|$}
		\FOR{$\alpha \in $}
		\STATE Build a DiFF-RF, $f$, using $X_i=\cup_{j\neq i} P_j$, $t$, $\psi$, $\alpha$;\\
		\STATE Evaluate the piece-wise anomaly scores on $P_i$ ($pwas(P_i)$);\\
		\STATE Evaluate the piece-wise anomaly scores on $P_i$ ($pwas(X_i)$);\\
		\STATE $R(\alpha) \leftarrow$ $R(\alpha)+\delta_Q(pwas(P_i), pwas(X_i))$;\\
		\ENDFOR	
		\ENDFOR
		\ENDFOR
		\STATE $\forall \alpha \in S_\alpha$, $R(\alpha)=R(\alpha)/\#iter$ //(see Eq.\ref{eq:distribDist});\\
		\RETURN $argmin_\alpha R$;
	\end{algorithmic}
	\caption{Function $get\_alpha(X, t, \psi)$}
	\label{Algo:AlphaTuning}
\end{algorithm}

\begin{figure}[h!]
	\centering
	\includegraphics[width=0.45\textwidth]{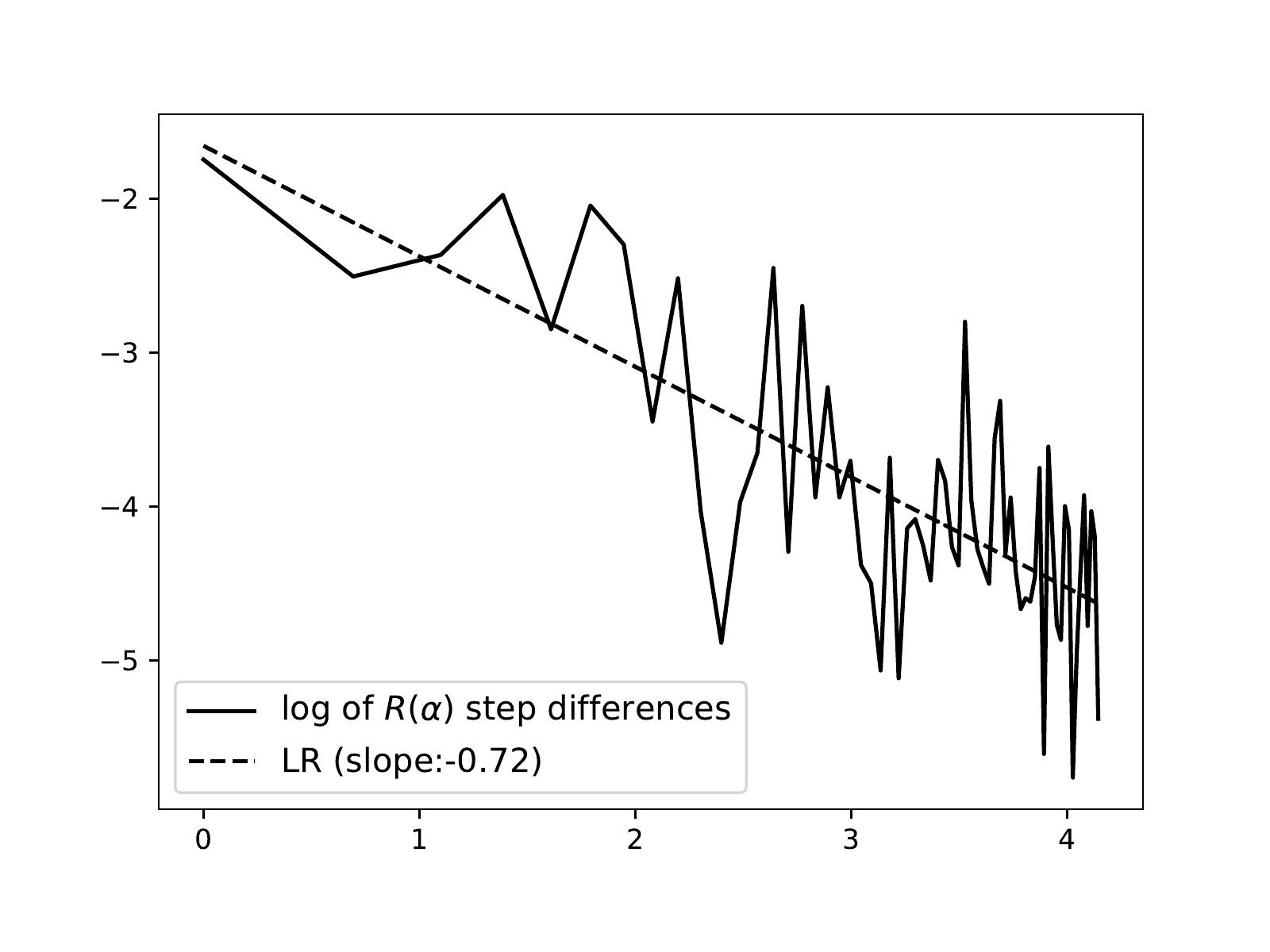}  
	\caption{$R(\alpha)$ step difference as a function of the iteration $k$ in logarithmic scale. In dotted line, the linear regression of the curve.}
	\label{fig:convergence}
\end{figure}

Figure \ref{fig:convergence} shows that the empirical convergence of the proposed cross-selection procedure is roughly $O(1/n)$ on the synthetic data. In practice we have observed similar convergence on all tested datasets. A theoretical statistical analysis could determine the conditions for the existence of an upper bound.

As a conclusion, this experiment tells us that $(t=128$, $\psi=.25\cdot|Xn|$) could be considered as a reasonable setting for small to medium size datasets. We adopt this configuration as the default setting for the DiFF-RF algorithm. $\alpha$ needs to be 'optimized' to best fit the data specificity. The procedure described in Algorithm \ref{Algo:AlphaTuning} can be used efficiently to achieve this goal as experimentally shown in the experimental section (Sec. \ref{sec:experimentation})

\subsection{Complexity of the DiFF-RF algorithm}
Basically, the DiFF-RF algorithm has the same overall complexity than the IF algorithm, although some extra computational costs are involved during training and testing stages.

IF has time complexities of $O(t\cdot\psi \cdot log(\psi))$ in the training stage and $O(n \cdot t \cdot log(\psi))$ in the testing stage.

\begin{figure}[h!]
	\centering
	\includegraphics[width=0.4\textwidth]{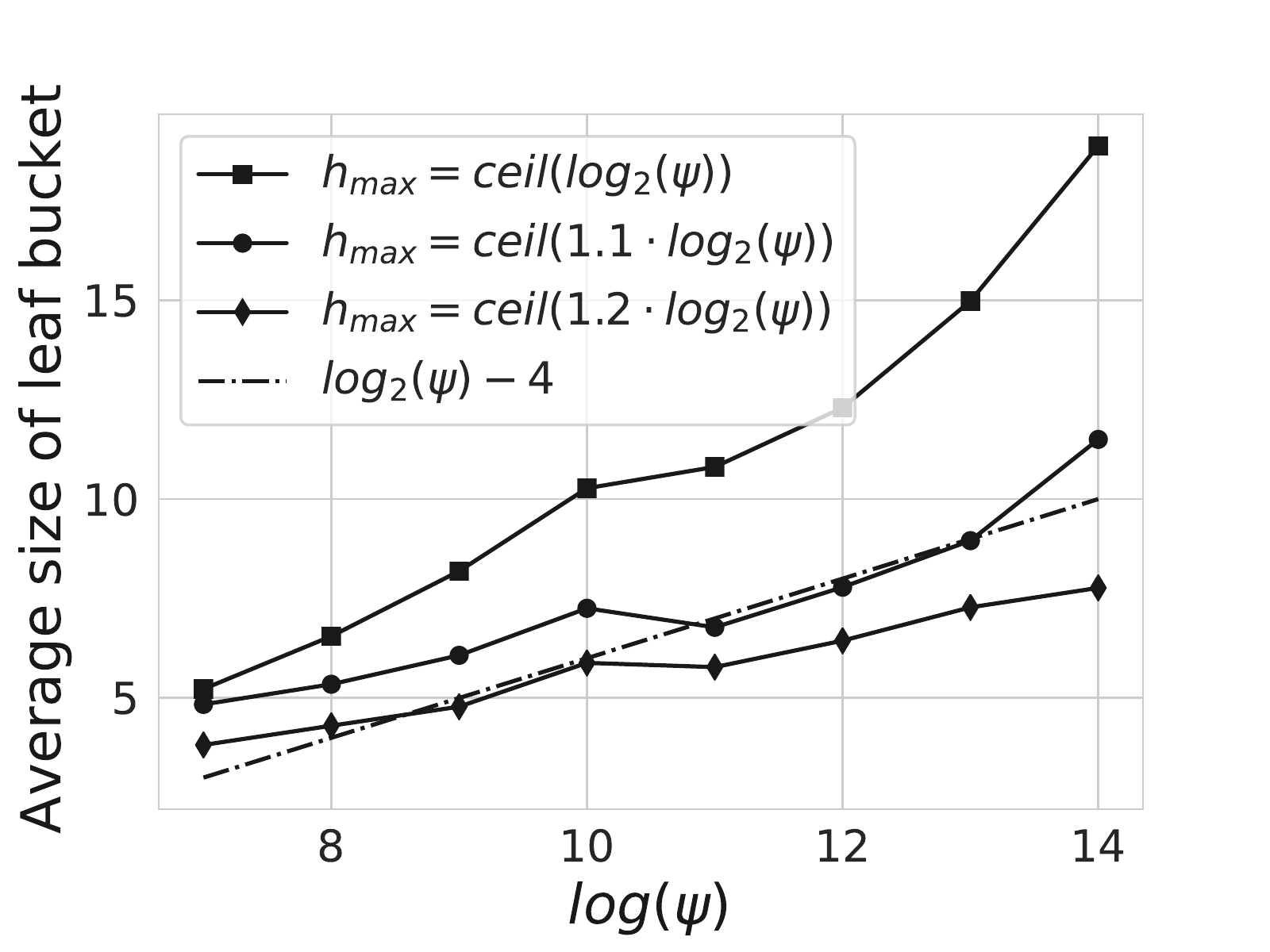}  
	\caption{Average number of instances contained in the external nodes of the iTrees as a function of $log_2(\psi)$; when the maximal height of the iTrees is $l_{max}=\ceil{log_2(\psi)}$ (dotted line),  $l_{max}=\ceil{1.1\cdot log_2(\psi)}$ (circle markers) and $l_{max}=\ceil{1.2\cdot log_2(\psi)}$ (square markers).}
	\label{fig:bucketSize}
\end{figure}

In addition, at training stage, the DiFF-RF algorithm requires to evaluate the centroids of the data attached to each of the leaf nodes, hence $\psi$ centroids in average need to be evaluated. The evaluation of a centroid is dependent on the number of elements contained in the leaf buckets ($n_{eb}$). It seems difficult to estimate precisely the expectation of $n_{eb}$, nevertheless, Fig.\ref{fig:bucketSize} presents the result of an empirical study that shows that if the maximal height of the DiFF tree is $h_{max}=\ceil{log_2(\psi)}$, then the average $n_{eb}$ value increases slightly faster than a logarithmic law. For this test, the random forest has been built from a normally distributed dataset with $(0.0,0.0)$ mean, and $((3, 0), (0, 3))$ covariance matrix. Hence, to maintain the overall algorithmic complexity at training stage close to $O(ln(\psi))$ one may increase slightly the maximal height of the DiFF tree. One can use for instance  $h_{max}=\ceil{1.2\cdot log_2(\psi)}$ that empirically maintains a sub-logarithmic growth as shown in Fig.\ref{fig:bucketSize}. At test time, the complexity of DiFF-RF is still $O(n \cdot t \cdot log(\psi))$ with a slight constant overhead compared to IF, due to the computation of the distances to leaf centroids.

\section{Experimentation}
\label{sec:experimentation}
We present below the results of the study we have carried out on some of the UCI's machine learning repository, supplemented by a study focusing on four intrusion detection benchmarks. We first describe the semi-supervised methods that we have evaluated in this comparative study and the datasets exploited to conduct our experiments.

\subsection{Evaluated anomaly detection models}
For this study, 7 semi-supervised models have been evaluated, basically a one-class SVM classifier (1C-SVM) \cite{Scholkopf2001}, a deep variational auto-encoder (VAE) \cite{Diederik2019}, an ensemble of auto-encoders (KitNET) \cite{Mirsky2018}, the isolation forest IF algorithm \cite{Liu2008} and its extended version EIF \cite{Hariri2018} and the DiFF-RF in its two modes, point-wise and collective anomaly detections.

For IF and DiFF-RF, the forests comprise $128$ trees and each tree is associated to a data sample containing $25\%$ of the training instances (with a maximum number set to $50k$ instances). 

We have used for IF and SVM the implementations provided by the SKLearn Python toolkit, and tensorflow with Keras for the VAE implementation. The VAE architecture is composed of symmetric encoder and decoder architectures implementing 6 dense layers and 3 drop-out layers. The latent space dimension has been setup to $10\%$ of the dimension of the original problem with a minimum of $3$ dimensions. The VAE has been trained using the \textit{Adam} optimizer.

The implementation of EIF that we have evaluated is freely available on github\footnote{https://github.com/sahandha/eif}. For IF and EIF, the default meta-parameters values have been used.

The tested KitNET implementation, also available on github\footnote{https://github.com/ymirsky/Kitsune-py}, does not require any meta-parameter setting.  
For the one class SVM, the default hyper-parameter values have been selected which is far to be optimal, but unfortunately none semi-supervised procedure is defined to tune the two hyper-parameters $\nu$ and $\gamma$ that are involved. 

For DiFF-RF (PW), the hyper-parameter $\alpha$ has been tuned on the training data according to the semi-supervised cross-selection procedure defined in subsection \ref{sec:meta_alpha}. The same $\alpha$ value has been used for DiFF-RF (CO).

\subsection{Heterogeneous domain datasets}
To assess the DiFF-RF algorithm in various domain area, we have selected 13 datasets in the UCI repository \cite{UCI} according to the following criteria: suitability for binary classification, multivariate numerical data and variability of the nature of the data (number of features, number of instances, distinct fields of application). A brief description of these datasets is given below.

\begin{enumerate}
	\item Banknote authentication (BNA): contains $1372$ instances described through $5$ features. The task consists to decide if a vectorized (real) representation of a banknote is legitimate or forgery. For our test, forgery data is considered as anomaly.
	\item Cardiotocography Data Set (CTG): contains 2126 instances described through 23 features. If the fetal state class code is normal (N), then the instance is considered as 'normal', otherwise it is considerd as an anomaly.
	\item Default of credit card clients (DCCC): contains $30000$ instances described through $24$ features. The task consists in predicting whether a credit card client will face payment default in the future. For our test, default data is considered as anomaly.
	\item HTRU2 \cite{Lyon2016}: contains $17898$ instances described through $9$ features. Each vector describes a sample of pulsar candidates collected during the High Time Resolution Universe Survey. For our test, the legitimate pulsar examples that belongs to the minority positive class, are considered as anomalies, and spurious examples, the majority negative class, are considered as normal data. 
	\item MAGIC Gamma Telescope Data (MAGIC): contains $19020$ instances described through $11$ features that characterize either primary gammas (majority class) signal or hadronic signal (minority class), considered for our test as anomaly.
	\item MiniBooNE particle identification (MiniBooNE): contains $130065$ instances described through $50$ features characterizing background signals, considered as the normal class, and the event signals, considered as the anomaly.
	\item MUSK (version 2): contains 6598 instances related to molecule conformations described through 168 features. Musk labeled conformations are considered as anomalies.
	\item Occupancy Detection (Occupancy) \cite{Candanedo2016}: contains $20560$ instances described through $7$ features that characterize the absence (normal) or presence (anomaly) of an individual in a room.
	\item Sensorless Drive Diagnosis Data Set (SDD): contains 58509 instances decribed through 49 attributes. Categories 1-9 are considered as 'normal' while categories 10-11 are considered as 'abnormal'.
	\item Spambase (SPAM): contains 4601 and 17720 instances described through 57 features. The task is to classify a vectorized (real) representation of a mail into normal or spam categories. For our test, spam data are considered as anomalies.
	\item Steel Plates Faults Data Set (SPF) \cite{Buscema2010}: contains 1941 instances with 27 attributes. Anomalies correspond to the presence of (at least) one of the 7 fault categories.
	\item TV News commercial detection \cite{Apoorv2014}, TVCD-BBC and TVCD-CNN: contain respectively 22535 and 22545 instances described through 4125 features. The task consists in detecting commercial spots in tv news: commercial spots are considered here as anomalies.
\end{enumerate}

In addition, we have created a \textit{Donut} dataset (DONUT-2.5) in a 5 dimensions space in which 2 dimensions represent the torus as defined in section \ref{donnut}, and 3 dimensions are $\mathcal{N}(0, .2)$ Gaussian noises.

\subsection{Intrusion detection datasets}
The initial motivation for this experiment is the detection of intrusion into networking systems, which requires consideration of the network's packet (or flow) data. To that end, four benchmark datasets are used to evaluate the approaches to be tested. 

\subsubsection{The ISCX dataset}
The ISCX dataset 2012 \cite{Shiravi:2012:TDS:2622690.2623143} has been prepared at the Information Security Centre of Excellence at the University of New Brunswick.
The entire ISCX labeled dataset comprises over two million traffic packets characterized with 20 features taking nominal, integer or float values. The dataset covers roughly seven days of network activities (i.e. normal and attack).
From this dataset, we have extracted 8 tasks according to the observed application protocol layer "HTTPWeb", "HTTPImageTransfer", "POP", "DNS", "SSH", "FTP", "SMTP", "ICMP".
Four different attack types, called as Brute Force SSH, Infiltrating, HTTP DoS, and DDoS are conducted on different days. 80\% of the normal data has been used as training, the remaining normal and attack data has been used as testing.


\subsubsection{The UNSW dataset}
The UNSW-NB15 dataset 2015 \cite{Moustafa2015} has been prepared at School of Engineering and IT, UNSW Canberra at ADFA, University of New South Wales.
The entire UNSW-NB15 labeled dataset comprises two million and 540,044 records which are stored in the four CSV files. Each record is described through 49 features taking nominal, integer or float values. From these files, we have extracted 6 tasks according to the application protocol layer  "HTTP", "FTP", "SMTP", "SSH", "DNS", "FTP-DATA".  
This data set contains nine types of attacks, namely, Fuzzers, Analysis, Backdoors, DoS, Exploits, Generic, Reconnaissance, Shellcode and Worms.
80\% of the normal data has been used as training, the remaining normal and attack data has been used as testing.

\subsubsection{The CIDDS dataset}
The CIDDS - benchmark \cite{Ring2017} has been prepared at HochSchule COBURG. It is composed of a series of flow based datasets specifically designed for anomaly-based network intrusion detection systems. We have used the CIDDS-1 OpenStack internal dataset, which comprises over 31 millions flows stored in CSV files. Each record is described through 19 features taking nominal, integer or float values. From these files, we have extracted 3 tasks according to the application transport layers  "ICMP", "TCP", "UDP".  This data contains Ping-Scans, Port-Scans, Brute-Force and Denial of  Service  attacks.

\subsubsection{The Kitsune dataset}
The Kitsune Network Attack Dataset \cite{Mirsky2018} has been developed at Ben-Gurion University of the Negev,Israel. The dataset is composed of 9 files covering 9 distinct attack (OS Scan, Fuzzing,  Video Injection, ARP Man in the Middle, Active Wiretap, SSDP Flood, SYN DoS, SSL Renegotiation, Mirai Botnet) situations. It is available for download at the UCI archive and comprises millions of records described through 115 features.

\subsection{Evaluation protocol}
For all the methods and datasets, we consider the Area Under the ROC curve (AUC) and the Average Precision (AP) as the evaluation measures. This avoids the need to set a threshold on the scoring values provided by the classifiers. 
For all tasks, 80\% of randomly selected normal data is used as training data, while the remaining 20\% is used for testing. All the attack/anomaly records are used for testing only.

\subsection{Results and discussion}

\begin{table*}[h!]
\centering
\caption{AUC and AP values obtained by AC-SVM, VAE and KitNET on the 40 datasets. \#train, \#test  stand respectively for the number of train, test data. (\#anomaly) is the total number of anomalies in the dataset.}
\label{tab:RES1}
\begin{tabular}{l|l|c|c|c|c}
\hline
&	&	1C-SVM	&	VAE	&	KitNET	&	\#train/\#test \\
Source &Dataset & AUC $|$ AP & AUC $|$ AP & AUC $|$ AP  &  (\#anomaly)\\ \hline\hline 	
&DONUT-2.5	&	 0.710 $|$ 0.763  	&	0.651 $|$ 0.818	&	0.736 $|$ 0.844	&	 2k/2.5k (1.5k)\\ \hline \hline 			
&BNA	&	0.963 $|$ 0.988	&	0.818 $|$ 0.945 	&	0.654 $|$ 0.903	&	 609/763 (610)\\  			
&CTG	&	0.858 $|$ 0.882	&	0.853 $|$ 0.894	&	0.889 $|$ 0.907	&	1324/802 (471)\\  			
&DCCC	&	0.475 $|$  0.555	&	 0.550 $|$ 0.627 	&	0.819 $|$ 0.901	&	 18K/11k (6.5k)\\  			
&HTRU2	&	0.826 $|$ 0.582	&	 0.944 $|$ 0.934	&	0.893 $|$ 0.882	&	 13K/5k (1.5k)\\  			
&MAGIC	&	0.684 $|$ 0.867	&	 0.705 $|$ 0.872	&	0.647 $|$ 0.863	&	 10K/9k (6k)\\  			
&MiniBooNE	&	0.790 $|$ 0.880	&	 0.820 $|$ 0.886 	&	0.642 $|$ 0.771	&	 75K/55k (36k)\\  			
UCI&MUSK	&	0.213 $|$ 0.328	&	0.267 $|$ 0.347	&	0.489 $|$ 0.483	&	4.5k/2k (1k)\\  			
&Occupancy	&	0.978 $|$ 0.973	&	 0.996 $|$   0.997	&	0.658 $|$ 0.713	&	 12.6K/7.9k (4.7k)\\  			
&SDD	&	0.501 $|$ 0.537	&	0.854 $|$ 0.872	&	1.000 $|$ 1.000	&	38k/20k (10.5k)\\ 			
&SPAM	&	0.642 $|$ 0.883	&	0.801 $|$ 0.900	&	0.755 $|$ 0.884	&	 2.2k/2.3k (1.8k)\\  			
&SPF	&	0.458 $|$ 0.684	&	0.439 $|$ 0.740	&	0.993 $|$ 0.995	&	1k/1k (673) \\ 			
&TVCD-BBC 	&	0.358 $|$ 0.763	&	 0.659 $|$ 0.900	&	0.680 $|$ 0.907	&	 7.4k/10.2k (8.2k)\\ 			
&TVCD-CNN	&	0.553 $|$ 0.911	&	 0.561 $|$ 0.903	&	0.682 $|$ 0.936	&	 6.5k/16k (14.4k)\\ \hline \hline 			
&HTTPWeb	&	0.927 $|$ 0.944	&	 0.932 $|$ 0.946 	&	0.936 $|$ 0.947	&	 40k/50k (40k)\\  			
&HTTPIT	&	0.595 $|$ 0.010	&	 0.567 $|$ 0.011 	&	0.600 $|$ 0.015	&	 40k/10k (64)\\  			
&POP	&	0.898 $|$ 0.198	&	 0.949 $|$ 0.769 	&	0.905 $|$ 0.702	&	 13k/3k (96) \\ 			
&IMAP	&	0.994 $|$ 0.843	&	0.994 $|$ 0.880	&	0.983 $|$ 0.680	&	 10k/3k (138) \\ 			
ISCX&DNS	&	0.496 $|$ 0.007	&	 0.821 $|$ 0.590 	&	0.823 $|$ 0.588	&	 40k/10k (73) \\ 		
&SSH	&	0.131 $|$ 0.834	&	0.980 $|$ 0.995	&	0.965 $|$ 0.987	&	 2k/10k (7,4k) \\ 			
&SMTP	&	0.643 $|$ 0.059	&	0.997 $|$ 0.922	&	0.998 $|$ 0.962	&	 7k/2k (76)\\ 			
&FTP	&	0.779 $|$ 0.254	&	0.998 $|$ 0.945	&	0.998 $|$ 0.949	&	 10k/2,5k (226)\\  			
&ICMP	&	0.348 $|$ 0.147	&	0.983 $|$ 0.787	&	0.997 $|$ 0.953	&	 6k/1.5k (295) \\ \hline \hline 			
&SSH	&	0.491 $|$ 0.002	&	1.000 $|$ 1.000	&	1.000 $|$ 1.000	&	37.5k/95k (19) \\  			
&FTP	&	0.505 $|$ 0.270	&	0.972 $|$ 0.843	&	0.812 $|$ 0.581	&	37k/12k (3k)\\ 			
UNSW&HTTP	&	0.509 $|$ 0.369	&	0.981 $|$ 0.940	&	0.975 $|$ 0.947	&	150k/55k (19k)\\ 			
&SMTP	&	0.514 $|$ 0.283	&	1.000 $|$ 1.000	&	1.000 $|$ 1.000	&	61k/20k (5k)\\ 			
&DNS	&	0.166 $|$ 0.487	&	1.000 $|$ 1.000	&	1.000 $|$ 1.000	&	460k/320k (210k)\\ \hline\hline	
&UDP	& 0.808 $|$ 0.867 &	0.695 $|$ 0.668&	0.940 $|$ 0.907& 8k/5.5k (3.3k)	 \\ 
CIDDS1&ICMP	&0.778 $|$ 0.137&	0.785 $|$ 0.051&	0.888 $|$ 0.094 & 5k/1.5k (20)	 \\ 
&TCP	&0.500 $|$ 0.100 &	0.775 $|$ 0.252&	0.720 $|$ 0.297&506k/145k (18k)	 \\ \hline\hline	
&Active\_Wiretap&	0.000 $|$ 0.000&	0.274 $|$ 0.651&	0.806 $|$ 0.884& 1084k/1194k (923k)	 \\ 
&ARP\_MitM	&0.000 $|$ 0.000&	 0.298 $|$ 0.699&	0.662 $|$ 0.853 & 1084k/1417k (1145k)	 \\ 
&Fuzzing	&0.000 $|$ 0.000&	0.494 $|$ 0.494&	0.829 $|$ 0.783& 1449/795k (433k)	 \\ 
&Mirai	&0.207 $|$ 0.945 &	 0.945 $|$ 0.998&	0.945 $|$ 0.998& 97k/667k (643k) 	 \\ 
KITSUNE&OS\_Scan	&0.000 $|$ 0.000&	0.900 $|$ 0.499& 	0.900 $|$ 0.503 & 1306k/392k (66k)	 \\ 
&SSDP\_Flood	&0.000 $|$ 0.000&	0.999 $|$ 0.999&	0.999 $|$ 0.999 & 2110k/1967k (528k)	\\ 
&SSL\_Renegotiation	&0.000 $|$ 0.000&	0.961 $|$ 0.727 &	0.983 $|$ 0.845 & 1692k/516k (93k)	 \\ 
&SYN\_DoS	&0.000 $|$ 0.000&	0.858 $|$ 0.268 &	0.723 $|$ 0.042	&  2211k/560k (7k)\\ 
&Video\_Injection	&0.000 $|$ 0.000&	0.595 $|$ 0.314&	0.952 $|$ 0.939& 1896k/576k (100k)	 \\ \hline\hline
&\textbf{Mean Rank}&	6.225 $|$ 6.2750 &	4.437 $|$ 4.2750 & 3.750 $|$  3.6875& -\\\hline
\end{tabular}
\end{table*}

\begin{table*}[h!]
	\centering
	\caption{AUC and AP values obtained by IF, EIF,DiFF-RF (point-wise) and DiFF-RF (collective) on the 40 datasets. The $\alpha$ meta parameter optimized on the training data is given in the last column.}
	\label{tab:RES2}
	\begin{tabular}{l|l|c|c|c|c|c}
    \hline										
Source &	Dataset	&	IF	&	EIF	&	 DiFF-RF (point-wise) 	&	 DiFF-RF (collective) 	&	alpha	\\\hline\hline
&	DONUT-2.5	&	0.679 $|$ 0.817	&	0.607 $|$ 0.761	&	0.961 $|$ 0.977	&	0.982 $|$ 0.990	&	20	\\
&	BNA	&	0.949 $|$ 0.984	&	0.969 $|$ 0.990	&	 1.000 $|$ 1.000	&	 1.000 $|$ 1.000	&	2	\\
&	CTG	&	0.823 $|$ 0.875	&	0.853 $|$ 0.884	&	0.809 $|$ 0.866	&	0.853 $|$ 0.899	&	1	\\
&	DCCC	&	0.536 $|$ 0.632	&	0.464 $|$ 0.561 	&	0.633 $|$ 0.696	&	0.684 $|$ 0.763	&	0.0001	\\
&	HTRU2	&	0.953 $|$ 0.943	&	0.939 $|$ 0.910	&	0.956 $|$ 0.949	&	0.952 $|$ 0.952	&	0.1	\\
&	MAGIC	&	0.818 $|$ 0.924	&	0.764 $|$ 0.903	&	0.853 $|$ 0.940	&	0.897 $|$ 0.956	&	1	\\
UCI &	MiniBooNE	&	0.744 $|$ 0.814	&	0.778 $|$ 0.839	&	0.757 $|$ 0.841	&	0.942 $|$ 0.965	&	0.05	\\
&	MUSK	&	0.362 $|$ 0.387	&	0.331 $|$ 0.378	&	0.569 $|$ 0.601	&	0.502 $|$ 0.553	&	0.5	\\
&	Occupancy	&	 0.947 $|$ 0.963	&	0.997 $|$ 0.998	&	0.900 $|$ 0.936	&	0.984 $|$ 0.993	&	0.01	\\
&	SDD	&	0.822 $|$ 0.840	&	0.777 $|$ 0.779	&	0.822 $|$ 0.859	&	0.901 $|$ 0.911	&	0.05	\\
&	SPAM	&	0.859 $|$ 0.954	&	0.855 $|$ 0.952	&	0.847 $|$ 0.932	&	0.883 $|$ 0.941	&	0.01	\\
&	SPF	&	0.415 $|$ 0.710	&	0.483 $|$ 0.743	&	0.656 $|$ 0.809	&	0.580 $|$ 0.771	&	0.5	\\
&	TVCD-BBC 	&	0.713 $|$ .0907	&	0.669 $|$ 0.885	&	 0.767 $|$ 0.925	&	0.804 $|$ 0.950	&	1	\\
&	TVCD-CNN	&	 0.593 $|$ 0.892	&	0.513 $|$ 0.892	&	 0.546 $|$ 0.903	&	0.813 $|$ 0.959	&	1	\\\hline\hline
&	HTTPWeb	&	 0.922 $|$ 0.953	&	0.905 $|$ 0.932	&	 0.962 $|$ 0.981	&	0.997 $|$ 0.998	&	0.1	\\
&	HTTPIT	&	 0.569 $|$ 0.013	&	0.588 $|$ 0.009	&	 0.605  $|$ 0.013	&	0.726 $|$ 0.021	&	1E-06	\\
&	POP	&	 0.936 $|$ 0.556	&	0.947 $|$ 0.749	&	 0.964 $|$ 0.921 	&	0.972 $|$ 0.926	&	0.1	\\
&	IMAP	&	0.995 $|$ 0.861	&	0.994 $|$ 0.800	&	0.998 $|$ 0.956	&	0.999 $|$ 0.974	&	0.1	\\
ISCX &	DNS	&	0.839 $|$ 0.545	&	0.833 $|$ 0.155	&	0.841 $|$ 0.598	&	0.844 $|$ 0.598	&	1	\\
&	SSH	&	0.995 $|$ 0.999	&	0.984 $|$ 0.996	&	0.997 $|$ 1.000	&	1.000 $|$ 1.000	&	0.0001	\\
&	SMTP	&	0.991 $|$ 0.688	&	0.987 $|$ 0.589	&	0.998 $|$ 0.958	&	0.999 $|$ 0.950	&	0.01	\\
&	FTP	&	0.995 $|$ 0.889	&	0.989 $|$ 0.777	&	0.997 $|$ 0.920	&	1.000 $|$ 0.997	&	0.001	\\
&	ICMP	&	0.982 $|$ 0.828	&	0.985 $|$ 0.865	&	0.996 $|$ 0.952	&	0.999 $|$ 0.980	&	0.1	\\\hline\hline
&	SSH\_UNSW	&	0.999 $|$ 0.686	&	0.997 $|$ 0.263	&	1.000 $|$ 1.000	&	1.000  $|$ 1.000	&	0.0001	\\
&	FTP\_UNSW	&	0.993 $|$ 0.973	&	0.979 $|$ 0.893	&	0.994 $|$ 0.981	&	0.996 $|$ 0.986	&	0.001	\\
UNSW &	HTTP\_UNSW	&	0.992 $|$ 0.972	&	0.968 $|$ 0.879	&	0.993 $|$ 0.978	&	0.995 $|$ 0.984	&	0.01	\\
&	SMTP\_UNSW	&	1.000 $|$ 1.000	&	1.000 $|$ 1.000	&	1.000 $|$ 1.000	&	1.000 $|$ 1.000	&	0.01	\\
&	DNS\_UNSW	&	0.999 $|$ 0.999	&	0.999 $|$ 0.998	&	1.000 $|$ 1.000	&	1.000 $|$ 1.000	&	0.01	\\\hline\hline
&	UDP\_CIDDS1	&	0.579 $|$ 0.006	&	 0.884 $|$ 0.023	&	0.338 $|$ 0.005	&	0.417 $|$ 0.005	&	0.100	\\
CIDDS &	ICMP\_CIDDS1	&	0.971 $|$ 0.999	&	0.981 $|$ 0.999	&	1.000 $|$ 1.000	&	1.000 $|$ 1.000	&	0.1	\\
&	TCP\_CIDDS1	&	0.998 $|$ 0.995	&	0.983 $|$ 0.928	&	1.000 $|$ 1.000	&	1.000 $|$ 1.000	&	1	\\\hline\hline
&	Active\_Wiretap	&	0.532 $|$ 0.729	&	0.496 $|$ 0.718	&	0.770 $|$ 0.904	&	0.845 $|$ 0.924	&	0.001	\\
&	ARP\_MitM	&	0.325 $|$ 0.719	&	0.280 $|$ 0.694	&	0.591 $|$ 0.845	&	0.486 $|$ 0.817	&	0.1	\\
&	Fuzzing	&	0.264 $|$ 0.431	&	0.332 $|$ 0.442	&	0.487 $|$ 0.547	&	0.995 $|$ 0.988	&	0.001	\\
Kitsune&	Mirai	&	0.953 $|$ 0.998	&	 0.893 $|$  0.989	&	0.991 $|$ 1.000	&	0.983 $|$ 0.999	&	0.01	\\
&	OS\_Scan	&	0.927 $|$ 0.578	&	0.897 $|$ 0.494	&	0.936 $|$ 0.625	&	0.936 $|$ 0.625	&	0.01	\\
&	SSDP\_Flood	&	0.999 $|$ 0.999	&	0.999 $|$ 0.997	&	0.999 $|$ 0.999	&	0.999 $|$ 0.999	&	0.1	\\
&	SSL\_Renegotiation	&	0.994 $|$ 0.955	&	0.973 $|$ 0.795	&	0.998 $|$ 0.992	&	0.968 $|$ 0.962	&	0.1	\\
&	SYN\_DoS	&	0.641 $|$ 0.246	&	0.624 $|$ 0.255	&	0.744 $|$ 0.243	&	0.796 $|$ 0.250	&	0.1	\\
&	Video\_Injection	&	0.801 $|$ 0.572	&	0.655 $|$ 0.297	&	0.947$|$ 0.910	&	0.905 $|$ 0.850	&	0.1	\\
 \hline\hline
&\textbf{Mean Rank}	&	4.175 $|$ 4.250 &	 4.737 $|$ 4.887&	2.775 $|$ 2.737 &	1.900 $|$  1.887& -\\ \hline
\end{tabular}
\end{table*}

We compare in Tables \ref{tab:RES1} and \ref{tab:RES2} the AUC and AP values obtained by the 7 benchmarked algorithms (1C-SVM, VAE, KitNet, IF, EIF, DiFF-RF (point-wise) and DiFF-RF (collective)), on 40 heterogeneous datasets, namely 13 tested UCI datasets, the donut synthetic data and the ISCX, UNSW, CIDDS and Kitsune intrusion detection benchmark datasets.

\begin{figure}[ht!]
	\centering
	\subfigure[]{
		\includegraphics[scale=0.50]{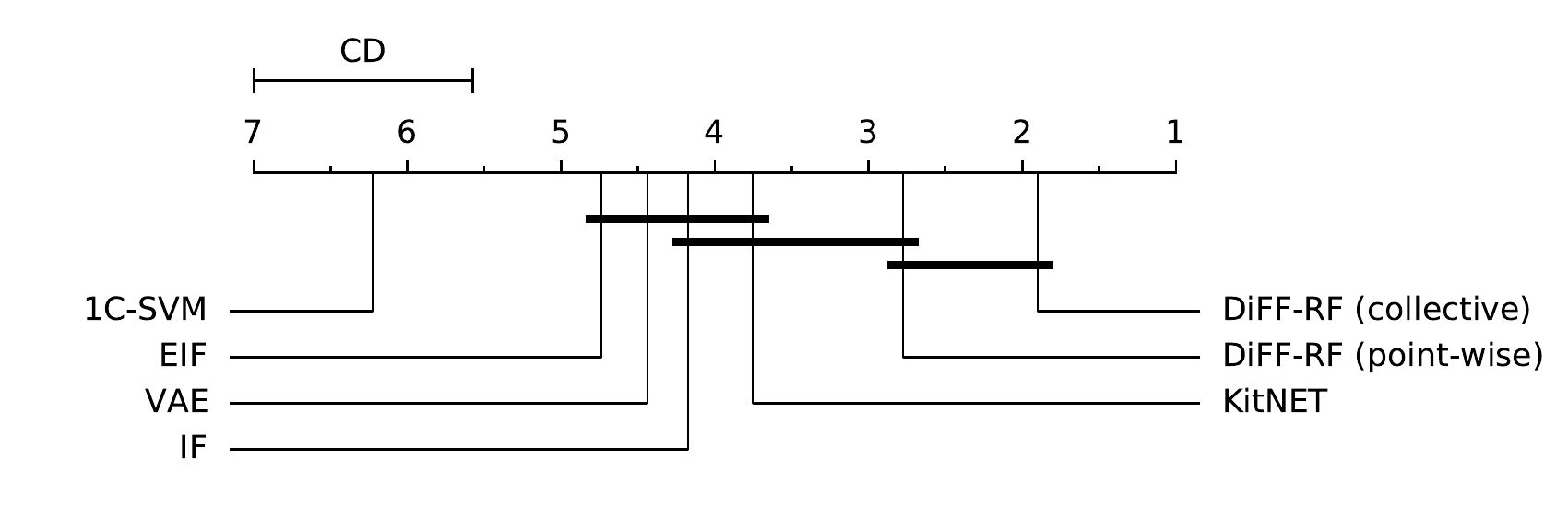} 
	}
	~
	\subfigure[]{
		\includegraphics[scale=0.50]{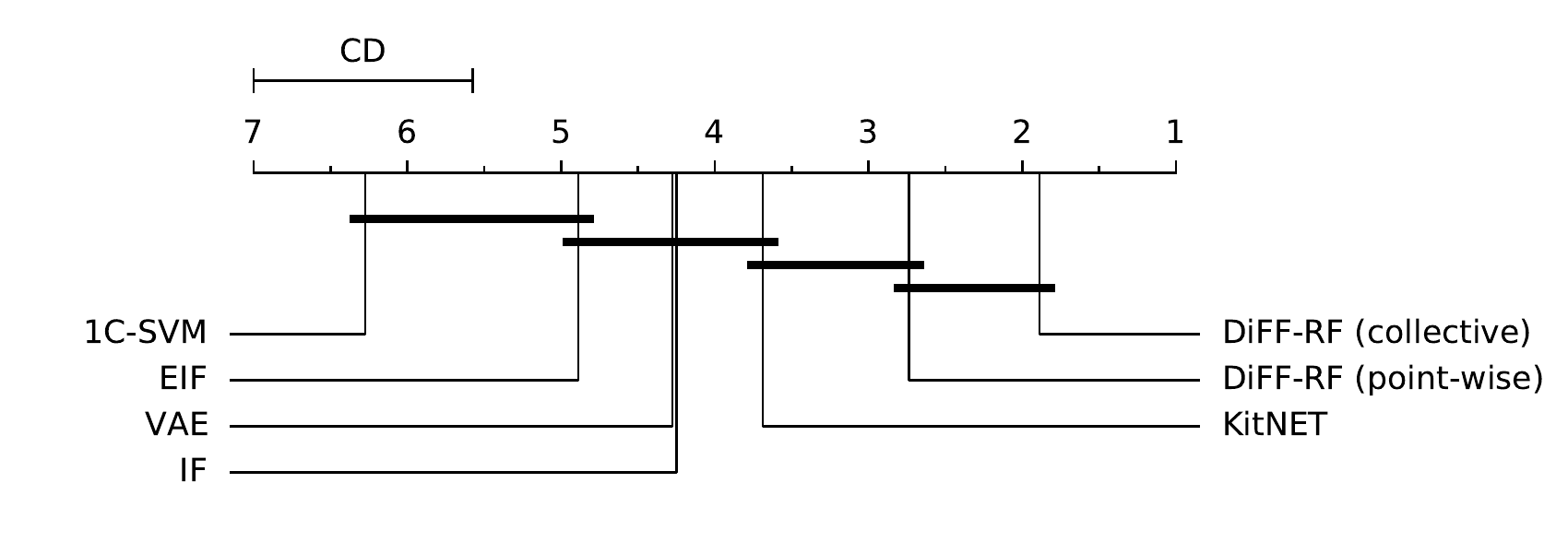} 
	}
	\caption{Ranking of the 7 classifiers with critical differences shown for the two evaluation measures: AUC (a) and AP (b)}
	\label{fig:criticalDiff}
\end{figure}

The last rows in Tables \ref{tab:RES1} and \ref{tab:RES2} gives the average rank for each method. DiFF-RF in its two configurations is the best ranked method in average. According to the AUC measure, collective DiFF-RF is ranked $2.19$, point-wise DiFF-RF $2.8$, KitNet $3.84$, IF $4.14$, VAE $4.32$, EIF $4.61$ and 1C-SVM  $6.1$. However, this overall result can give an over optimistic view of the situation, as, for some datasets, AUC or AP differences may not be always significant. To obtain a better synthetic overview of these results, we performed a Friedman's nonparametric test to determine whether there are significant differences between the tested classifiers. We use the post-hoc Nemenyi test to infer which differences are significant. The result of this statistical analysis is given in figures \ref{fig:criticalDiff}-(a) for AUC and \ref{fig:criticalDiff}-(b) for AP measures respectively. Based on the post-hoc Nemenyi test, and according to the AUC measure, we assume that there are no significant differences within the following groups: \{DiFF-RF (collective), DiFF-RF (point-wise)\}; \{DiFF-RF (point-wise), KitNET\} ; \{KitNET, VAE, IF, and EIF\}; \{1C-SVM\}. All the other differences are considered as significant.

In light of these results, the following observations can be made:
\begin{itemize}
	\item the one class SVM is not doing well, not only because it does not scale well and does not provide a response for large datasets in a reasonable elapsed time, but also because of the difficulty to select the meta parameters (default values were used, and this not a good guess in general).
	\item EIF performs slightly worst than IF in average although the two algorithms fall in the same group. Noticeably, EIF performs poorly on the DONUT dataset, showing that the `blind spot' problem still exists for EIF.
	\item Point-wise DiFF-RF outperforms significantly the IF and EIF algorithms, showing that the `blind spot' effect could play a role in some applications. The greatest performance gaps are observed for the DONUT, BNA, CTG MUSK, SPAM, HTTPWeb-ISCX, HTTPIT-ISCX, ICMP-ISCX, UDP\_CIDDS-1, ARP\_MitM, Fuzzing, SYN\_DoS and Video\_Injection datasets.
	\item  Point-wise DiFF-RF achieves significantly better results in average compared to the deep variational auto-encoder implementation  although, in some cases such as MiniBooNE or SPAM, VAE performs much better than DiFF-RF. 
	\item Point-wise DiFF-RF falls in the same group than KitNET but ranks better in average than this deep ensemble architecture. However, KitNET outperforms greatly DiFF-RF (PW) on three among the nine Kitsune datasets (for which the KitNET architecture has been precisely designed): Active\_Wiretap, ARP\_MitM and Fuzzing. Furthermore, it fails comparatively to Kitnet on several datasets such as UDP\_CIDDS1 or DCCC.
	\item In its collective anomaly detection configuration, DiFF-RF is particularly efficient and outperforms significantly all the other approaches. In some rare cases, such as for the UDP\_CIDDS1 datasets, it gets lower AUC and AP values than the other methods. One may notice however, that for these datasets, the detection tasks is quite difficult, and all the methods performed poorly (AP values below $.02$). 
	\item Although the evaluated datasets are not specifically designed for the purpose of collective anomaly detection, except maybe for the intrusion detection data that contains some Deny of Service (DoS) attacks, the implementation of the frequency of visit criteria seems to be quite effective to characterize abnormal co-occurrences of events that can be, if taken separately, considered as normal.
	\item On the intrusion detection tasks, the DiFF-RF implementations are particularly efficient, except for the HTTPIT-ISCX,Active\_Wiretap, ARP\_MitM and Fuzzing datasets. As other methods perform poorly on these data, one can incriminate the lost of information when encoding the payload during the pre-procesing step. The number of features describing image data for instance is obviously inadequate. 
	\item Finally, the semi-supervised cross-selection procedure defined and used to tune the single extra hyper-parameter ($\alpha$) that is introduced in DiFF-RF seems to be adequate.
\end{itemize}

Notice that in our experiment, it should be noted that method DiFF-RF in its collective anomaly configuration has the advantage of having simultaneous knowledge of all the test data (excepted their labels). This supplemental information was not available for the other methods. \\

\begin{table}[h!]
	\centering
	\caption{Elapsed computation time (in sec.) spent by the 5 tested classifiers (1C-SVM is much too long to be tested).}
	\begin{tabular}{|l|c|c|c|c|c|}
	\hline
	 & VAE & KitNET & IF & EIF & DiFF-RF \\
	Dataset    &     &        &    &     &(co \& pw)\\\hline\hline
	UDPCIDDS1 & 2776 & 1900 & 71 & 1184 & 1644\\
	SSDP\_Flood & 2017 &2515 &243 &799 & 3647\\\hline
	\end{tabular}
	\label{tab:ET}
\end{table}

The elapsed computational time obtained on the two largest data sets, UDPCDDIS (23 features, 3.4M training instances)) and Kitsune-SSDP\_Flood (115 features, 2M training instances), spent by the tested classifiers is given in Table \ref{tab:ET}. We clearly see that DiFF-RF, that is not parallelized during testing time, is slower than the IF classifier. However, on the UDPCIDDS dataset, its elapsed processing time has the same order of magnitude than VAE and KiTNET. It degrades a bit on the SSDP\_Flood dataset, mainly because the algorithmic complexity of DiFF-RF increases linearly with the dimension of the data, due to the distance computations. The IF SkLearn implementation that is highly parallelized is between 1 and 2 order of magnitude faster than DiFF-RF, showing that there is room for improvement regarding the implementation of DiFF-RF.

\section{How can DiFF-RF help to harden IDS?}

No intrusion detection method is perfect. As our experiments show, the top-ranked tested methods, DiFF-RF and Kitsune, are rather complementary: on some datasets, Kitsune works well while DiFF-RF does not, and conversely. On most datasets, their performances are not significantly different. On the other hand, the detection models on which they are built are significantly different (forest of random partitioning trees for DiFF-RF and set of auto-encoders for Kitsune). Therefore, the appropriate combination of these ensemble methods could be considered in order to develop a more robust meta-detector, more resilient to concept drift, while ensuring that the design of adverse attacks is more difficult. We develop below a case to support the use of DiFF-RF as a contribution to the hardening IDS from an operational point of view.

\subsection{Integrability}
DiFF-RF is a software component that can be easily integrated into an SIEM by processing collected samples of flows or packets from a probe placed on a network to provide alarms in case of unexpected traffic. However, the implementation of the algorithm as it currently stands, can be optimized if an online use is considered. Some engineering work can be done to distribute the evaluation of the trees on the available cores of the host machine to reach a one order of magnitude in speed-up on common modern hardware. Some effort can also be spent to reduce the memory occupancy. The integration of the component into an IDS processing pipeline such as SNORT should not raised any major difficulties in practice.
DiFF-RF can currently be used as a forensic analysis tool with some efficiency to locate abnormal data in sliding windows of flows or packets. Collective anomaly scoring is particularly effective for detecting distributed or automated attacks such as DoS, Flood or Fuzzing attacks.

\subsection{Resistance to concept drift}
\label{subsec:drift}
Learning under concept drift, that impacts specifically the processing of streaming data, in particular IDS, is a challenge well reviewed in \cite{Lu2019}.
Concept drift is formally defined as the change in joint distribution of a set of input variables X and a target variable y. In a semi-supervised learning framework, the concept drift that we consider is simply the change of the distribution of X representing the 'legitimate' data.\\

The collective anomaly scoring implemented in DiFF-RF seems particularly well suited to identify drift by jointly examining over some time windows how distances to leaf centroids and frequencies of visits into tree leaves evolve. The challenge is indeed to separate the legitimate drifts from the attacks patterns. But one have first to establish the ability of the detector to effectively detect changes in the covariate distribution P(X).

\begin{figure}[ht!]
	\centering
	\subfigure[]{
		\includegraphics[scale=0.15]{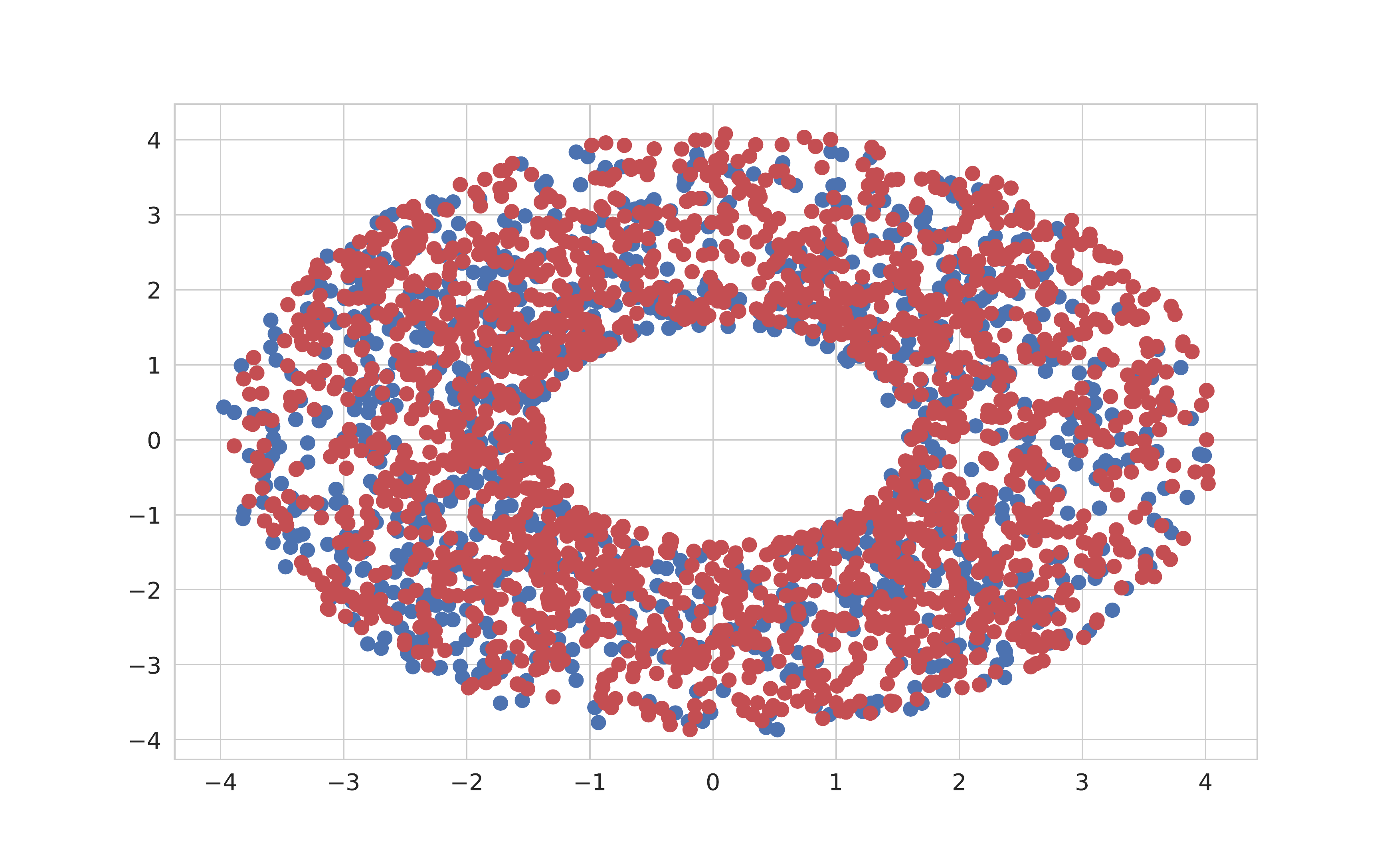} 
	}
	~
	\subfigure[]{
		\includegraphics[scale=0.25]{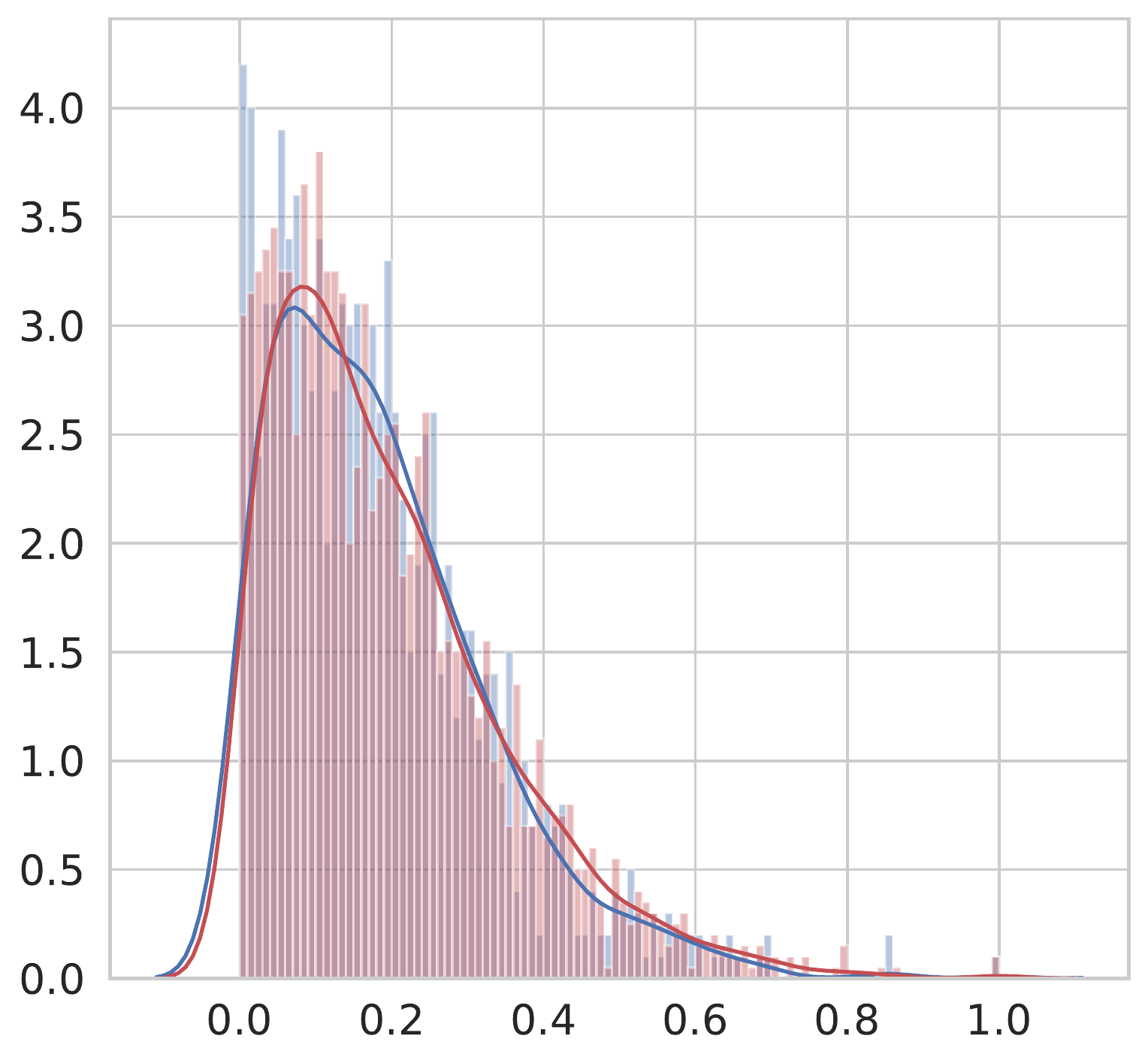} 
	}
	~
\subfigure[]{
	\includegraphics[scale=0.25]{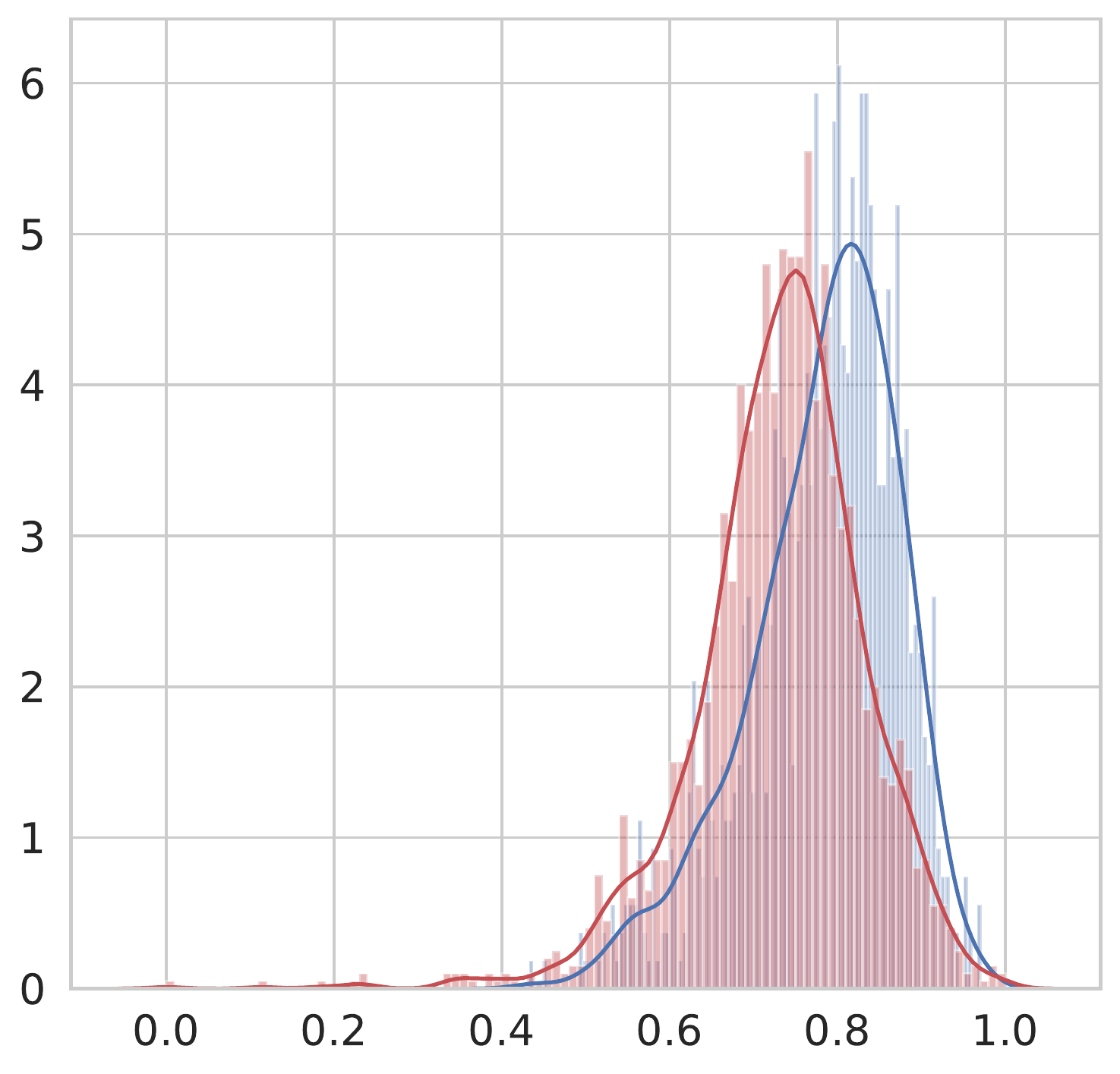} 
}
	\caption{(a) Translation drift in Donut data, (b) KitNet score distributions and (c) DiFF-RF collective anomaly score distributions (blue before drift, red after drift)}
	\label{fig:Tdrift}
\end{figure}

The following synthetic tests show that the collective anomaly score provided by the DiFF-RF reacts even in the presence of small global drifts in normal training data. In this experiment, a 4\% translation drift was first applied to the Donut data set, as shown in figure \ref{fig:Tdrift} on the left. The center and right sub-figures show the distributions obtained by the KitNet and DiFF-RF (collective anomaly score) detectors respectively.  For all three sub-figures, the initial distribution of the normal data is shown in blue, while the distribution associated with the drifted data is shown in red. Two homothetic drifts have also been proposed to complete this study. Table \ref{tab:drift} shows that, according to the two-sample Kolmogorov-Smirnov test, the distribution corresponding to the DiFF-RF collective anomaly score of the normal data is detected as being significantly different from that obtained from the drifted data. Note that this is not the case for the DiFF-RF point-wise score or the score provided by KitNet for which both distributions are considered to belong to the same random variable.

A variance increase or decrease would be similarly detected by the collective anomaly score of DiFF-RF as shown for the homothetic drifts in Table \ref{tab:drift}, and undetected by KitNet or DiFF-RF point-wise.
Hence one may expect that DiFF-RF can alert on some global concept drifts resulting in  (statistically significant) changes in the empirical distributions of its anomaly scores. These drifts still need to be analyzed by an expert to verify that none attack is underway. However, this type of check does not need to be done online or frequently if the concept drift is slow, and a re-training of the DiFF-RF could then be done to cope with the traffic drift.  

\begin{table}[h!]
	\centering
	\caption{Two-sample Kolmogorov–Smirnov test for the translation and homothetic drifts on Donnuts data. Critical statistic value at alpha=.01, alpha=.005 and alpha=.001 are respectively 0.073, 0.077 and 0.087.}
	\begin{tabular}{|l|c|c|}
		\hline
		 \textbf{Translation drift (4\%)} & statistic & p-value \\\hline\hline
		KitNet  & 0.033  & 0.439 \\
		DiFF-RF (point-wise) & 0.052 & 0.032 \\
		DiFF-RF (collective)  & \textbf{0.283} & \textbf{2.22e-15} \\\hline
		\textbf{homothetic drift (+4\%)} & statistic & p-value \\\hline\hline
		KitNet & 0.046 & 0.118\\
		DiFF-RF (point-wise) & 0.035 & 0.478 \\
		DiFF-RF (collective)  & \textbf{0.287} & \textbf{2.22e-15}\\\hline
		\textbf{Homothetic drift (-4\%)} & statistic & p-value \\\hline\hline
		KitNet  & 0.024 & 0.834 \\
		DiFF-RF (point-wise) & 0.033 & 0.479 \\
		DiFF-RF (collective)  &\textbf{ 0.294} & \textbf{2.22e-15}\\\hline	
	\end{tabular}
	\label{tab:drift}
\end{table}	

An adversarial attack experiment is proposed in the next section based on a flooding of legitimate patterns at the vicinity of a malicious pattern. It is a Deny Of Service attempt against the detector based on a local crafted drift. This last study complements the global drift experiments proposed above.

\subsection{Mitigating the impact of some adversarial attacks}

Adversarial attacks to semi-supervised IDS fall mostly into three broad categories:
\begin{enumerate}
	\item Poisoning attacks: for such kind of attacks, the adversary has access jointly to the training data and to the training process. The attacker modifies the training data by adding some poisoning data in the neighborhood of a malicious pattern (nothing forbids to add directly into the training data some malicious patterns) and is able to re-train the detection model with the poisonous crafted data before putting it back to exploitation.   
	\item Evasion attacks: these attacks are in general crafted from known malicious patterns that are correctly detected by the IDS. The malicious pattern is `minimally' modified until reaching a new similar and still malicious pattern that will go undetected through the IDS. Such attacks need as background information the knowledge of the trained detection model, that the attacker can use at least as a black box.
	\item Flooding attacks: such attacks use large amount of legitimate data to target areas near the decision boundary of the detector, with the aim at increasing drastically the False Positive Rate, hence making the detector unusable.  \\
\end{enumerate}

In a semi-supervised intrusion detection framework, poisoning attacks are difficult to detect. One can expect that the manipulation of the training data by the attacker will result to a sort of concept drift. In the absence of an un-poisoned normal data distribution that can be used as a reference, it is quite impossible to detect a malicious engineering of the training data, although some attempts exist for near neighbor based detector such as DBSCAN or Local Outlier Factor detection \cite{Bhargava2019}. Conversely, if a reference distribution exists for normal data, and if a global or local drift is attempted, DiFF-RF would be able to detect it, as far as the drifts have significant (statistical) impact on the frequencies of visit in the leaf nodes of the trees corresponding to the areas where the drifts are performed, as shown previously in subsection \ref{subsec:drift}. \\

Evasion attacks can indeed succeed if and only if a malicious pattern can be crafted in the very close neighborhood of normal data. If we accept the double-hypothesis that no normal data is malicious and that the feature space is a correct embedding in the scope of the detection task, this means that the new malicious pattern should be located in a so-called `blindspot' of the detection model. Identifying and removing `blind spots' is hence a natural approach to fight against adversarial attacks. DiFF-RF has been precisely designed with this orientation while removing an important `blind spot' in the IF algorithm. The point-wise scoring of DiFF-RF  is based on the randomized sampling of the local densities in which the normal observations are embedded. If normal training data is available to correctly estimate local densities everywhere, then no `blind spots' could exist in theory. 

However, in real life, available training data is not infinite, such data is potentially characterized in high dimension feature space, the data embedding is not perfect and some discriminating features may be missing, memory is limited and response time is an issue. This means that estimating local densities of normal data everywhere in the feature space is difficult in practice and we have to accept the presence of "noise" in our estimations that would naturally result in local `blind spots' that could be exploited. 


One can argue that, due to the randomization of the partition trees, local `blindspots' in DiFF-RF, generated by the lack of training data or missing relevant features, are randomly spread throughout the whole forest. In practice, an attacker will need to have full access to the trained forest to attempt, if possible, to identify some of such `blind spots' not too far from a malicious pattern. 
Finally, the attacker still has to ensure that his attack  does not significantly alter the frequency of visit at leaf level. To avoid this somehow unlikely scenario, we could recommend to re-train periodically the DiFF-RF, to cope with potential concept drift and limit as well the risk for an attacker to get the trained model and the trained data. \\

\begin{figure}[ht!]
	\centering
	\includegraphics[scale=0.9]{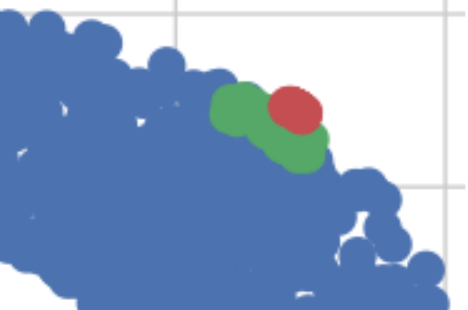} 
	\caption{Flooding experiment carried on the synthetic Donuts data. blue: legit training data, green: flooding legit data, red: attack data}
	\label{fig:floodExperiment}
\end{figure}

\begin{table}[ht!]
	\caption{\label{tab:floodResults}Evaluation of the flood attack.}
	\centering
	\setlength\extrarowheight{2pt}
	\begin{tabular}{|l|l|l|l|l|} 
		\cline{2-5}
		\multicolumn{1}{l|}{}    & \multicolumn{2}{l|}{\begin{tabular}[c]{@{}l@{}}DiFF-RF\\point-wise\end{tabular}} & \multicolumn{2}{l|}{\begin{tabular}[c]{@{}l@{}}DiFF-RF\\collective\end{tabular}}  \\ 
		\cline{2-5}
		\multicolumn{1}{l|}{}   & AUC  & EER   & AUC  & EER   \\ 
		\hline
		\begin{tabular}[c]{@{}l@{}}red as attacks\\ no green data\end{tabular}  & 0.939 & 0.086  & 1.0 & 0.0  \\ 
		\hline
		\begin{tabular}[c]{@{}l@{}}red as attacks\\ green legit\end{tabular}  & 0.856 & 0.169  & 1.0 & 0.007  \\ 
		\hline
		\begin{tabular}[c]{@{}l@{}}red and green~\\as attacks\end{tabular} & 0.889 & 0.161  & 1.0 & 0.001  \\
		\hline
	\end{tabular}
\end{table}

Flooding attacks will be easily detected by DiFF-RF since they are expected to alter the frequency of visit in the leaves of the trees that cover the vicinity of targeted spots. The collective anomaly detection is particularly well suited to identify such situation as shown by the example depicted in Fig. \ref{fig:floodExperiment}. In this experiment, the red spot represents 10 attack data and the green spot 50 legitimate data used to perform the flood attack. 1000 legitimate data are randomly drawn from the global normal data distribution to simulate a normal 'traffic'. Note that the (red) attack cluster has an intersection with the legitimate data to make the detection task more difficult.

Table \ref{tab:floodResults} gives the AUC measure and the Equal Error Rate (EER), i.e the error rate evaluated when False Positives equal False Negatives, for three setups: i)basic case, no flooding data is used, ii) flooding data is added and considered as normal data and iii) flooding data is added and considered as attack data. \\

The experiment clearly shows that the flooding succeeds for the point-wise detection since the AUC drops from .939 to .856 (EER increases from .086 to .169) if we consider flooding data as normal or from AUC=.939 to AUC=.889 (EER increases from .086 to .161) if we consider flooding data as attack data. However, the experiment shows that the flooding has almost no impact on the collective anomaly detection. In other words, the flooding do not managed to fool the frequency of visit in the leaf nodes of the DiFF-RF trees.


\section{Conclusion}

We have introduced the semi-supervised DiFF-RF algorithm dedicated to anomaly detection. DiFF-RF is an ensemble approach based on random partitioning trees. It comes with two configurations depending on whether one considers point-wise or collective anomalies. From the construction of a synthetic dataset, thanks to a distance criteria introduced at the leaf level of the partitioning trees, we have shown that DiFF-RF solves an apparently quite penalizing drawback observed in the Isolation Forest algorithm, the so-called `blindspots' effect, that characterizes unoccupied areas in the data embedding as 'normal' areas even if they are far from the 'normal' data distribution.

Furthermore, considering frequency of visit at leaf level DiFF-RF provides this algorithm with the ability to cope with collective anomaly very efficiently while significantly improving in general the scores obtained by the point-wise configuration.

Extensive testing on UCI datasets and on four benchmarks dedicated to intrusion detection,  shows that DiFF-RF is quite efficient at detecting point-wise or so-considered collective anomalies, comparatively to the state of the art methods in this domain, namely ensemble of deep auto-encoders, single variational auto-encoder, Isolation Forest, the Extended Isolation Forest and One-Class Support Vector Machine. 

However, DiFF-RF is failing to provide satisfactory detection on some datasets. The effects of the o and should be further analyzed to to get some insights about its misdetection.

Furthermore, similarly to the IF algorithm, DiFF-RF scales relatively well comparatively to One-Class SVM and VAE and parallelized implementations are obviously possible. 

Our experimentation shows also that the proposed semi-supervised cross-selection of the extra hyper-parameter that is introduced in DiFF-RF algorithm to scale the distance calculation at leaf level is well suited in practice.

Indeed, more experimentation should be carried out using  datasets dedicated to collective anomaly detection to explore the limits of the DiFF-RF approach on this task.  However, as it stands, DiFF-RF is a quite competitive semi-supervised approach for anomaly and intrusion detection with some resilience to concept drift and adversarial flooding attacks.

One straightforward perspective is to evaluate and adapt the DiFF-RF algorithm for processing streaming data. The use of a sliding window along the stream would allow for better investigating the collective anomaly detection capability through the detail analysis of the variations in the relative frequency of visit at leaf levels.
Another perspective is to extend DiFF-RF to cope with categorical data in addition to numerical data. That would require to implement dedicated similarity or distance measure for categorical data, while replacing the mean calculation at leaf level by the selection of a medoid for instance. The frequency of visit at leaf level criteria would remain unchanged.
Finally, the use of a clustering algorithm at leaf level would allow for coping with situations where several residual clusters of data samples are coexisting at the leaves.

The DiFF-RF Python implementation is available at \url{https://github.com/pfmarteau/DiFF-RF}.

%
%

\bibliographystyle{IEEEtran}
\bibliography{IEEEabrv, biblio}

\begin{thebibliography}{10}
\providecommand{\url}[1]{#1}
\csname url@samestyle\endcsname
\providecommand{\newblock}{\relax}
\providecommand{\bibinfo}[2]{#2}
\providecommand{\BIBentrySTDinterwordspacing}{\spaceskip=0pt\relax}
\providecommand{\BIBentryALTinterwordstretchfactor}{4}
\providecommand{\BIBentryALTinterwordspacing}{\spaceskip=\fontdimen2\font plus
\BIBentryALTinterwordstretchfactor\fontdimen3\font minus
  \fontdimen4\font\relax}
\providecommand{\BIBforeignlanguage}[2]{{%
\expandafter\ifx\csname l@#1\endcsname\relax
\typeout{** WARNING: IEEEtran.bst: No hyphenation pattern has been}%
\typeout{** loaded for the language `#1'. Using the pattern for}%
\typeout{** the default language instead.}%
\else
\language=\csname l@#1\endcsname
\fi
#2}}
\providecommand{\BIBdecl}{\relax}
\BIBdecl

\bibitem{Agrawal2015}
\BIBentryALTinterwordspacing
S.~Agrawal and J.~Agrawal, ``Survey on anomaly detection using data mining
  techniques,'' \emph{Procedia Computer Science}, vol.~60, pp. 708 -- 713,
  2015. [Online]. Available:
  \url{http://www.sciencedirect.com/science/article/pii/S1877050915023479}
\BIBentrySTDinterwordspacing

\bibitem{Fitriani2016}
S.~{Fitriani}, S.~{Mandala}, and M.~A. {Murti}, ``Review of semi-supervised
  method for intrusion detection system,'' in \emph{2016 Asia Pacific
  Conference on Multimedia and Broadcasting (APMediaCast)}, 2016, pp. 36--41.

\bibitem{Khraisat2019}
A.~Khraisat, I.~Gondal, P.~Vamplew, and J.~Kamruzzaman, ``Survey of intrusion
  detection systems: techniques, datasets and challenges.''
  \emph{Cybersecurity}, vol.~2, no.~1, pp. 1--20, 2019.

\bibitem{Chandola:2009}
\BIBentryALTinterwordspacing
V.~Chandola, A.~Banerjee, and V.~Kumar, ``Anomaly detection: A survey,''
  \emph{ACM Comput. Surv.}, vol.~41, no.~3, pp. 15:1--15:58, Jul. 2009.
  [Online]. Available: \url{http://doi.acm.org/10.1145/1541880.1541882}
\BIBentrySTDinterwordspacing

\bibitem{Goldstein2016}
M.~Goldstein and S.~Uchida, ``A comparative evaluation of unsupervised anomaly
  detection algorithms for multivariate data,'' \emph{PLOS ONE}, vol.~11,
  no.~4, pp. 1--31, 04 2016.

\bibitem{Mennatallah2012}
M.~Amer and M.~Goldstein, ``Nearest-neighbor and clustering based anomaly
  detection algorithms for rapidminer,'' in \emph{Proceedings of the 3rd
  RapidMiner Community Meeting and Conferernce (RCOMM 2012)}, S.~Fischer and
  I.~Mierswa, Eds.\hskip 1em plus 0.5em minus 0.4em\relax Shaker Verlag GmbH, 8
  2012, pp. 1--12.

\bibitem{key:articleBama}
S.~Bama, M.~Ahmed, and A.Saravanan, ``Article: Network intrusion detection
  using clustering: A data mining approach,'' \emph{International Journal of
  Computer Applications}, vol.~30, no.~4, pp. 14--17, September 2011.

\bibitem{Lin201513}
\BIBentryALTinterwordspacing
W.-C. Lin, S.-W. Ke, and C.-F. Tsai, ``Cann: An intrusion detection system
  based on combining cluster centers and nearest neighbors,''
  \emph{Knowledge-Based Systems}, vol.~78, pp. 13 -- 21, 2015. [Online].
  Available:
  \url{http://www.sciencedirect.com/science/article/pii/S0950705115000167}
\BIBentrySTDinterwordspacing

\bibitem{Xiong2011}
L.~Xiong, B.~P\'{o}czos, and J.~Schneider, ``Group anomaly detection using
  flexible genre models,'' in \emph{Proceedings of the 24th International
  Conference on Neural Information Processing Systems}, ser. NIPS’11.\hskip
  1em plus 0.5em minus 0.4em\relax Red Hook, NY, USA: Curran Associates Inc.,
  2011, p. 1071–1079.

\bibitem{Desir2013}
\BIBentryALTinterwordspacing
C.~D{\'e}sir, S.~Bernard, C.~Petitjean, and L.~Heutte, ``One class random
  forests,'' \emph{Pattern Recogn.}, vol.~46, no.~12, pp. 3490--3506, Dec.
  2013. [Online]. Available:
  \url{http://dx.doi.org/10.1016/j.patcog.2013.05.022}
\BIBentrySTDinterwordspacing

\bibitem{Fujimaki2008}
R.~Fujimaki, ``Anomaly detection support vector machine and its application to
  fault diagnosis,'' in \emph{2008 Eighth IEEE International Conference on Data
  Mining}, Dec 2008, pp. 797--802.

\bibitem{Lee:1998:DMA:1267549.1267555}
\BIBentryALTinterwordspacing
H.~Debar, M.~Becker, and D.~Siboni, ``A neural network component for an
  intrusion detection system,'' in \emph{Proceedings 1992 IEEE Computer Society
  Symposium on Research in Security and Privacy}.\hskip 1em plus 0.5em minus
  0.4em\relax Berkeley, CA, USA: USENIX Association, 1992, pp. 240--250.
  [Online]. Available: \url{http://dl.acm.org/citation.cfm?id=1267549.1267555}
\BIBentrySTDinterwordspacing

\bibitem{Ghosh:1999:SUN:1251421.1251433}
\BIBentryALTinterwordspacing
A.~K. Ghosh and A.~Schwartzbard, ``A study in using neural networks for anomaly
  and misuse detection,'' in \emph{Proceedings of the 8th Conference on USENIX
  Security Symposium - Volume 8}, ser. SSYM'99.\hskip 1em plus 0.5em minus
  0.4em\relax Berkeley, CA, USA: USENIX Association, 1999, pp. 12--12.
  [Online]. Available: \url{http://dl.acm.org/citation.cfm?id=1251421.1251433}
\BIBentrySTDinterwordspacing

\bibitem{ryan:nips10}
\BIBentryALTinterwordspacing
J.~Ryan, M.-J. Lin, and R.~Miikkulainen, ``Intrusion detection with neural
  networks,'' in \emph{Advances in Neural Information Processing Systems 10},
  M.~I. Jordan, M.~J. Kearns, and S.~A. Solla, Eds.\hskip 1em plus 0.5em minus
  0.4em\relax Cambridge, MA: MIT Press, 1998, pp. 943--949. [Online].
  Available: \url{http://nn.cs.utexas.edu/?ryan:nips97}
\BIBentrySTDinterwordspacing

\bibitem{Kemmler2013}
\BIBentryALTinterwordspacing
M.~Kemmler, E.~Rodner, E.-S. Wacker, and J.~Denzler, ``One-class classification
  with gaussian processes,'' \emph{Pattern Recognition}, vol.~46, no.~12, pp.
  3507 -- 3518, 2013. [Online]. Available:
  \url{http://www.sciencedirect.com/science/article/pii/S0031320313002574}
\BIBentrySTDinterwordspacing

\bibitem{Lee2001}
\BIBentryALTinterwordspacing
W.~Lee and D.~Xiang, ``Information-theoretic measures for anomaly detection,''
  in \emph{Proceedings of the 2001 IEEE Symposium on Security and Privacy},
  ser. SP '01.\hskip 1em plus 0.5em minus 0.4em\relax Washington, DC, USA: IEEE
  Computer Society, 2001, pp. 130--. [Online]. Available:
  \url{http://dl.acm.org/citation.cfm?id=882495.884435}
\BIBentrySTDinterwordspacing

\bibitem{GFDLS2006}
G.~Gu, P.~Fogla, D.~Dagon, W.~Lee, and B.~\v{S}kori\'{c}, ``Towards an
  information-theoretic framework for analyzing intrusion detection systems,''
  in \emph{11th European Symposium on Research in Computer Security (ESORICS),
  Hamburg}.\hskip 1em plus 0.5em minus 0.4em\relax Springer, Sep 2006, pp.
  527--546, lNCS Vol. 4189.

\bibitem{Akoglu2015}
\BIBentryALTinterwordspacing
L.~Akoglu, H.~Tong, and D.~Koutra, ``Graph based anomaly detection and
  description: A survey,'' \emph{Data Min. Knowl. Discov.}, vol.~29, no.~3, pp.
  626--688, May 2015. [Online]. Available:
  \url{http://dx.doi.org/10.1007/s10618-014-0365-y}
\BIBentrySTDinterwordspacing

\bibitem{Kramer1991}
\BIBentryALTinterwordspacing
M.~A. Kramer, ``Nonlinear principal component analysis using autoassociative
  neural networks,'' \emph{AIChE Journal}, vol.~37, no.~2, pp. 233--243, 1991.
  [Online]. Available:
  \url{https://aiche.onlinelibrary.wiley.com/doi/abs/10.1002/aic.690370209}
\BIBentrySTDinterwordspacing

\bibitem{Vincent2010}
P.~Vincent, H.~Larochelle, I.~Lajoie, Y.~Bengio, and P.-A. Manzagol, ``Stacked
  denoising autoencoders: Learning useful representations in a deep network
  with a local denoising criterion,'' \emph{J. Mach. Learn. Res.}, vol.~11, p.
  3371–3408, Dec. 2010.

\bibitem{Diederik2019}
\BIBentryALTinterwordspacing
D.~P. Kingma and M.~Welling, ``An introduction to variational autoencoders,''
  \emph{Foundations and Trends® in Machine Learning}, vol.~12, no.~4, pp.
  307--392, 2019. [Online]. Available:
  \url{http://dx.doi.org/10.1561/2200000056}
\BIBentrySTDinterwordspacing

\bibitem{Mirsky2018}
M.~Yisroel, D.~Tomer, E.~Yuval, and S.~Asaf, ``Kitsune: An ensemble of
  autoencoders for online network intrusion detection,'' in \emph{Network and
  Distributed System Security Symposium 2018 (NDSS'18)}, 2018.

\bibitem{Liu2008}
F.~T. Liu, K.~M. Ting, and Z.~H. Zhou, ``Isolation forest,'' in
  \emph{Proceedings of the 8th IEEE International Conference on Data Mining
  (ICDM'08)}, 2008, pp. 413--422.

\bibitem{Liu2012}
\BIBentryALTinterwordspacing
F.~T. Liu, K.~M. Ting, and Z.-H. Zhou, ``Isolation-based anomaly detection,''
  \emph{ACM Trans. Knowl. Discov. Data}, vol.~6, no.~1, pp. 3:1--3:39, march
  2012. [Online]. Available: \url{http://doi.acm.org/10.1145/2133360.2133363}
\BIBentrySTDinterwordspacing

\bibitem{Ding2013}
\BIBentryALTinterwordspacing
Z.~Ding and M.~Fei, ``An anomaly detection approach based on isolation forest
  algorithm for streaming data using sliding window,'' \emph{IFAC Proceedings
  Volumes}, vol.~46, no.~20, pp. 12 -- 17, 2013. [Online]. Available:
  \url{http://www.sciencedirect.com/science/article/pii/S1474667016314999}
\BIBentrySTDinterwordspacing

\bibitem{Aryal2014}
\BIBentryALTinterwordspacing
Sunil, K.~Ting, J.~Wells, and T.~Washio,
  ``\BIBforeignlanguage{English}{Improving iforest with relative mass},'' in
  \emph{\BIBforeignlanguage{English}{Advances in Knowledge Discovery and Data
  Mining: 18th Pacific-Asia Conference, Proceedings (PAKDD 2014), Part II}},
  V.~Tseng, T.~Ho, Z.-H. Zhou, A.~Chen, and H.-Y. Kao, Eds.\hskip 1em plus
  0.5em minus 0.4em\relax Springer, 2014, pp. 510 -- 521, pacific-Asia
  Conference on Knowledge Discovery and Data Mining 2014, PAKDD 2014 ;
  Conference date: 13-05-2014 Through 16-05-2014. [Online]. Available:
  \url{https://sites.google.com/site/pakdd2014/,
  https://link.springer.com/book/10.1007/978-3-319-06608-0}
\BIBentrySTDinterwordspacing

\bibitem{Shen2016}
Y.~Shen, H.~Liu, Y.~Wang, Z.~Chen, and G.~Sun, \emph{A Novel Isolation-Based
  Outlier Detection Method}.\hskip 1em plus 0.5em minus 0.4em\relax Cham:
  Springer International Publishing, 2016, pp. 446--456.

\bibitem{Liao2018}
L.~L. and L.~B., ``Entropy isolation forest based on dimension entropy for
  anomaly detection,'' in \emph{Computational Intelligence and Intelligent
  Systems (ISICA 2018)}, vol. 986.\hskip 1em plus 0.5em minus 0.4em\relax
  Springer, 2018.

\bibitem{Cheng19}
\BIBentryALTinterwordspacing
Z.~Cheng, C.~Zou, and J.~Dong, ``Outlier detection using isolation forest and
  local outlier factor,'' in \emph{Proceedings of the Conference on Research in
  Adaptive and Convergent Systems}, ser. RACS '19.\hskip 1em plus 0.5em minus
  0.4em\relax New York, NY, USA: Association for Computing Machinery, 2019, p.
  161–168. [Online]. Available: \url{https://doi.org/10.1145/3338840.3355641}
\BIBentrySTDinterwordspacing

\bibitem{Hariri2018}
\BIBentryALTinterwordspacing
S.~Hariri, M.~C. Kind, and R.~J. Brunner, ``Extended isolation forest,''
  \emph{CoRR}, vol. abs/1811.02141, 2018. [Online]. Available:
  \url{http://arxiv.org/abs/1811.02141}
\BIBentrySTDinterwordspacing

\bibitem{Marteau2017}
\BIBentryALTinterwordspacing
P.-F. Marteau, S.~Soheily-Khah, and N.~B{\'e}chet, ``{Hybrid Isolation Forest -
  Application to Intrusion Detection},'' May 2017, working paper or preprint.
  [Online]. Available: \url{https://hal.archives-ouvertes.fr/hal-01520720}
\BIBentrySTDinterwordspacing

\bibitem{Scholkopf2001}
B.~Sch{\"o}lkopf, J.~Platt, J.~Shawe-Taylor, A.~Smola, and R.~Williamson,
  ``Estimating the support of a high-dimensional distribution.'' \emph{Neural
  Computation}, vol.~13, no.~7, pp. 1443--1471, Mar. 2001.

\bibitem{UCI}
\BIBentryALTinterwordspacing
D.~Dua and C.~Graff, ``{UCI} machine learning repository,'' 2017. [Online].
  Available: \url{http://archive.ics.uci.edu/ml}
\BIBentrySTDinterwordspacing

\bibitem{Lyon2016}
\BIBentryALTinterwordspacing
R.~J. Lyon, B.~W. Stappers, S.~Cooper, J.~M. Brooke, and J.~D. Knowles,
  ``{Fifty years of pulsar candidate selection: from simple filters to a new
  principled real-time classification approach},'' \emph{Monthly Notices of the
  Royal Astronomical Society}, vol. 459, no.~1, pp. 1104--1123, 04 2016.
  [Online]. Available: \url{https://doi.org/10.1093/mnras/stw656}
\BIBentrySTDinterwordspacing

\bibitem{Candanedo2016}
\BIBentryALTinterwordspacing
L.~M. Candanedo and V.~Feldheim, ``Accurate occupancy detection of an office
  room from light, temperature, humidity and co2 measurements using statistical
  learning models,'' \emph{Energy and Buildings}, vol. 112, pp. 28--39, jan
  2016. [Online]. Available:
  \url{https://doi.org/10.1016\%2Fj.enbuild.2015.11.071}
\BIBentrySTDinterwordspacing

\bibitem{Buscema2010}
M.~{Buscema}, S.~{Terzi}, and W.~{Tastle}, ``A new meta-classifier,'' in
  \emph{2010 Annual Meeting of the North American Fuzzy Information Processing
  Society}, 2010, pp. 1--7.

\bibitem{Apoorv2014}
\BIBentryALTinterwordspacing
A.~Vyas, R.~Kannao, V.~Bhargava, and P.~Guha, ``Commercial block detection in
  broadcast news videos,'' in \emph{Proceedings of the 2014 Indian Conference
  on Computer Vision Graphics and Image Processing}, ser. ICVGIP ’14.\hskip
  1em plus 0.5em minus 0.4em\relax New York, NY, USA: Association for Computing
  Machinery, 2014. [Online]. Available:
  \url{https://doi.org/10.1145/2683483.2683546}
\BIBentrySTDinterwordspacing

\bibitem{Shiravi:2012:TDS:2622690.2623143}
\BIBentryALTinterwordspacing
A.~Shiravi, H.~Shiravi, M.~Tavallaee, and A.~A. Ghorbani, ``Toward developing a
  systematic approach to generate benchmark datasets for intrusion detection,''
  \emph{Computer Security}, vol.~31, no.~3, pp. 357--374, May 2012. [Online].
  Available: \url{http://dx.doi.org/10.1016/j.cose.2011.12.012}
\BIBentrySTDinterwordspacing

\bibitem{Moustafa2015}
N.~Moustafa and J.~Slay, ``Unsw-nb15: a comprehensive data set for network
  intrusion detection systems (unsw-nb15 network data set),'' 11 2015.

\bibitem{Ring2017}
M.~Ring, S.~Wunderlich, D.~Grüdl, D.~Landes, and A.~Hotho, ``Creation of
  flow-based data sets for intrusion detection,'' \emph{Journal of Information
  Warfare}, vol.~16, pp. 40--53, 2017.

\bibitem{Lu2019}
J.~{Lu}, A.~{Liu}, F.~{Dong}, F.~{Gu}, J.~{Gama}, and G.~{Zhang}, ``Learning
  under concept drift: A review,'' \emph{IEEE Transactions on Knowledge and
  Data Engineering}, vol.~31, no.~12, pp. 2346--2363, 2019.

\bibitem{Bhargava2019}
R.~Bhargava, ``Adversarial anomaly detection,'' Ph.D. dissertation, Purdue
  University, 8 2019.

\end{thebibliography}


\end{document}